\documentclass{article}

\usepackage{microtype}
\usepackage{graphicx}
\usepackage{subfigure}
\usepackage{booktabs} % for professional tables
\usepackage{enumitem}

\usepackage{hyperref}

\usepackage[accepted]{icml2023}

\usepackage{amsfonts}
\usepackage{amsmath}
\usepackage{amssymb}
\usepackage{mathtools}
\usepackage{amsthm}

\usepackage[capitalize,noabbrev]{cleveref}

\theoremstyle{plain}

\theoremstyle{definition}

\theoremstyle{remark}

\usepackage[textsize=tiny]{todonotes}

\icmltitlerunning{Chameleon: Adapting to Peer Images for Planting Durable Backdoors in Federated Learning}

\begin{document}

\twocolumn[
\icmltitle{Chameleon: Adapting to Peer Images \\ 
    for Planting Durable Backdoors in Federated Learning}



\icmlsetsymbol{equal}{*}

\begin{icmlauthorlist}
\icmlauthor{Yanbo Dai}{ustgz}
\icmlauthor{Songze Li}{ustgz,ust}
\end{icmlauthorlist}

\icmlaffiliation{ustgz}{IoT Thrust, Information Hub, Hong Kong University of Science and Technology (Guangzhou), Guangzhou, China}
\icmlaffiliation{ust}{Department of Computer Science and Engineering, Hong Kong University of Science and Technology, Hong Kong SAR, China}

\icmlcorrespondingauthor{Songze Li}{songzeli@ust.hk}

\icmlkeywords{Federated Learning, Backdoor Attack, Contrastive Learning, ICML}

\vskip 0.3in
]



\printAffiliationsAndNotice{}  % leave blank if no need to mention equal contribution

% \begin{abstract}
%     Due to clients having full control over their dataset and training process, federated learning (FL) systems are especially vulnerable to adversarial backdoor attacks. Existing methods have proven that it is feasible to insert visual backdoors into FL systems. However, backdoors planted by these methods may not be durable and are easy to vanish once adversaries leave the training procedure. In this paper, we discover that the durability of visual backdoors can be greatly enhanced by utilizing sample relationships. Further, we discover two novel factors which affects the durability of backdoors in FL and propose a new method, Chameleon, based on that. We also provide an extensive empirical evaluation on multiple datasets to illustrate Chameleon outperforms prior methods on durability under all the evaluated settings.
% \end{abstract}

\begin{abstract}
    In a federated learning (FL) system, distributed clients upload their local models to a central server to aggregate into a global model. Malicious clients may plant backdoors into the global model through uploading poisoned local models, causing images with specific patterns to be misclassified into some target labels. Backdoors planted by current attacks are not durable, and vanish quickly once the attackers stop model poisoning. In this paper, we investigate the connection between the durability of FL backdoors and the relationships between benign images and poisoned images (i.e., the images whose labels are flipped to the target label during local training). Specifically, benign images with the original and the target labels of the poisoned images are found to have key effects on backdoor durability. Consequently, we propose a novel attack, Chameleon, which utilizes contrastive learning to further amplify such effects towards a more durable backdoor. Extensive experiments demonstrate that Chameleon significantly extends the backdoor lifespan over baselines by $1.2\times \sim 4\times$, for a wide range of image datasets, backdoor types, and model architectures. 
    % based on that. We also provide an extensive empirical evaluation on multiple datasets to illustrate Chameleon outperforms prior methods on durability under all the evaluated settings.
\end{abstract}

\vspace{-7mm}
\section{Introduction}
\label{intro}
Rocketing development in computational resources brings exponential growth in both personal and corporate data. To sufficiently exploit these data and break the data isolation between decentralized data owners, collaborative training frameworks that respect data privacy and resist malicious participants are needed \cite{lyu2022privacy}.
%the training paradigm needs to be both privacy-preserving and robust to malicious behaviors. 
Federated learning (FL) \cite{mcmahan2017communication}, consisting of a group of data-owning clients collaborating through a central server, is an emerging collaborative training paradigm, in which clients do not share their raw data with either the server or other clients, but participate in training a global model through uploading local models or gradients to the server. While protecting clients' data privacy, FL systems are vulnerable to a number of malicious attacks \cite{bagdasaryan01, Zhengming01, fang2020local, geiping2020inverting, fung2020limitations, shejwalkar2022back, nasr2019comprehensive, bhagoji2019analyzing}, which hampers FL from being used in real-world settings.

% Attackers can perform model poisoning attacks to damage the performance of the global model by uploading the local malicious model whose parameters are deliberately manipulated. 

Malicious clients can launch model poisoning attacks to damage the performance of the global model, via uploading deliberately manipulated local models. According to attacker's goals, model poisoning attacks can be categorized into untargeted attacks \cite{fang2020local, shejwalkar2021manipulating, chen2017distributed, guerraoui2018hidden} and backdoor attacks \cite{bhagoji2019analyzing, bagdasaryan01}. Compared with untargeted attacks whose goal is to compromise the overall performance of the global model, backdoor attacks are especially destructive because of its strong stealth. Specifically, backdoor attackers try to have the global model misclassify to a particular target label when it encounters certain image patterns (or triggers), and perform normally on other inference tasks without triggers.

% other inference tasks when it does not. 

% predict the particular target label instead of the original label when it encounters certain image patterns and does not interfere with other inference tasks when it does not. 

Previous works \cite{bagdasaryan01,Zhengming01,Hongyi01,zitengsun01,Yuxin01} have successfully planted backdoors in FL models. However, these works usually need attackers to keep uploading poisoned models to avoid planted backdoors from vanishing. This requires attackers to consistently participate in the training process, which will make the embedded backdoor easier to be detected. It has been observed in current attacks \cite{bagdasaryan01, Zhengming01} that once all attackers leave the training procedure or the global model is retrained after malicious attacks stop, backdoors will quickly be erased by model updates from benign clients\cite{xie2021crfl}. In such cases, attackers wish to plant more durable backdoors to ensure that the inserted backdoors survive until the FL system is deployed. 

As for the reason for rapidly vanishing backdoors, it is still not fully understood. Most previous works \cite{Zhengming01, Yuxin01} try to explain from the perspective of model conflicts.
%update conflict of local models from the benign clients. 
They argue that the poisoned model trained by the malicious client conflicts with benign models on certain key parameters. When attackers no longer send the poisoned model to the central server, following uploaded benign models will gradually erase the malicious client's contribution on these parameters, and thus backdoor will vanish. \cite{Hongyi01} proposes edge-case attacks from the perspective of backdoor samples' distribution. Also, they train the poisoned model using projected gradient descent, which periodically projects the model parameters on a ball centered around the model of the previous iteration, to escape the norm-clipping defense that mitigates the effect from abnormally large updates. They argue that backdoors trained by data samples that live in the tail of the input distribution are more durable and thus resistant to erasing. Because these edge samples have less chance to conflict with other samples.

In this work, we conduct first-principle investigations on the key factors that affect backdoor durability in FL, through the lens of relationships between sample images,
%from the perspective of peer images 
and propose a new backdoor attack, Chameleon, which allows attackers to plant more durable visual backdoors \textit{through adapting to peer images}. We observe that the durability of backdoors is mainly dominated by the presence of two types of peer benign images that are closely related to the poisoned images: 1) \textit{interferers}: images that share the same original labels with the poisoned images; and 2) \textit{facilitators}: images with target backdoor label. Interferers are much likely to cause update conflicts between the poisoned updates and benign updates, which may cause the drop of backdoor accuracy. Facilitators, on the other hand, can help to reintroduce the backdoor information to the FL model, and slow down the catastrophic forgetting effect after adversaries leave the training process. Motivated by this observation, we design Chameleon to amplify these effects on the backdoor durability.
% Updates submitted by clients possessing more facilitators can help to retain the backdoor accuracy. The motivation of focusing on peer images is that we believe that different samples have different effect on the inserted backdoors. Thus, it is essential to identify which samples have the highest impact on the backdoor durability, and consider how to effectively exploit these impact to enhance the backdoor durability.

% two key factors which affect the durability of backdoors: first is the update conflicts caused by benign images that share the same original label with the poisoned images 
% interferers and the slow down of catastrophic forgetting by facilitators. 
% The key insight of Chameleon is about how to treat the relationship between poisoned images and other images properly. 
Specifically, the local training process at a corrupted client is divided into two stages. In the first stage, contrastive learning is employed to train a representation encoder, such that the embedding distance between poisoned images and the interferers is pushed further, and the embedding distance between poisoned images and the facilitators is pulled closer.
% makes poisoned images adapt to peer images. 
% Through the first training stage, the embedding distance between poisoned images and images with the target label, which we call \textit{facilitators}, is pulled closer while that between poisoned images and images with the original label, which we call \textit{interferers}, is pushed further. 
Without changing the network architecture, the second stage of Chameleon simply trains a classifier while fixing the parameters of the encoder obtained in the first stage.
% Then, we freeze the parameters of the representation encoder and train a classifier based on that in the second stage. Through adopting the novel contrastive learning paradigm driven by the observations, our new poisoned model is more durable and stealthy as we note that the model architecture will not be changed through the training process.

We provide extensive empirical evaluations on multiple computer vision datasets, including CIFAR10, CIFAR100, and EMNIST, with both pixel-pattern backdoors and semantic backdoors \cite{bagdasaryan01} against a FL system equipped with norm-clipping defense \cite{zitengsun01}. We show that Chameleon outperforms all the prior backdoor attacks in durability for both fixed and dynamic pattern backdoors, by increasing the backdoor lifespan by $1.2\times \sim 4\times$. Further, we also evaluate the performance of different backdoor attacks under different model architectures. We find that Chameleon is less sensitive to the change of model architecture and achieves strong backdoor durability consistently. In sharp contrast, backdoors inserted by prior methods fail to persist and fades quickly for deeper networks.

\section{Related Work}
\label{relatedwork}
\textbf{Federated Averaging}. FedAVG \cite{mcmahan2017communication} is the baseline algorithm for implementing FL systems. Generally speaking, FedAVG aims to minimize the summation of the local empirical losses $\sum_{i=1}^n F_i(\theta)$ of $n$ participating clients, over a global model $\theta$ in a decentralized manner. For every global round $t$, the server broadcasts the current global model $\theta^t$ to a subset of randomly selected clients. Each selected client then optimizes its local loss $F_i(\theta)$ over local dataset, and uploads obtained local model $\theta_i^t$ back to the server.
% and the received parameter $w^t$ for a certain number of local rounds. Client $i$ then uploads the local model parameter $w_i^t$ back to the server. 
The central server then performs model aggregation to update the model $\theta^{t+1}=\frac{1}{n}\sum_{i=1}^n \theta_i^t$. In the rest of the paper, we will use FedAVG as our method to implement an FL system.

\textbf{Backdoor Attacks}. Backdoor attacks are initially proposed to poison deep learning models \cite{gu2019badnets, chen2017targeted, liu2017trojaning, liu2020reflection, shafahi2018poison, saha2020hidden}. BadNets~\cite{gu2019badnets} first implements the backdoor attack by inserting pixel-pattern backdoors into deep learning models. In order to escape human examination on the malicious training dataset, \cite{chen2017targeted} further proposes an invisible attack method that generates backdoor images by blending the backdoor trigger with benign images. \cite{liu2017trojaning} explores optimized backdoor attacks which generate backdoor triggers through solving optimization problems. \cite{bagdasaryan01} proposes the first backdoor attack targeting FL systems by uploading models with large norms to replace the original global model with the poisoned model. DBA \cite{xie2019dba} utilizes the decentralized nature of FL and achieves better performance by decomposing a global trigger pattern into separate local trigger patterns which are embedded into training dataset of multiple malicious parties.

\textbf{Defenses}. %There are a number of recently proposed defense methods that can effectively mitigate various backdoor attacks. 
According to \cite{rieger2022close}, defenses against backoor attacks in FL can mainly be categorized into influence reduction approaches that constrain the impact of individual updates \cite{zitengsun01, andreina2021baffle, naseri2022local, yin2018byzantine} and detection and filtering approaches that aim to identify and exclude poisoned updates \cite{nguyen2022flame, rieger2022deepsight, shen2016auror, zhao2020shielding}.
Among all the defense approaches, the norm-clipping defense proposed in~\cite{zitengsun01} is the most widely adopted method. It regularizes the norm of each model update received at the server within a norm bound $\rho$, which was shown to be especially effective in mitigating model replacement backdoors \cite{bagdasaryan01}.
%which needs to upload models with irregularly large parameters fails under the defense of norm-clipping. Norm-clipping tries to regularize the norm of the model updates received by the server within the agreed norm bound $C$. 
%FLAME \cite{nguyen2022flame} takes a step further by additionally requiring the server to check the cosine similarity among all the received gradients. The server will first cluster all the uploaded gradients based on cosine similarity, and then mark outliers as malicious updates and exclude them from the aggregation procedure. Finally, the server add gaussian noise to make all selected gradients differentially private \cite{abadi2016deep}. 
There are also some general Byzantine-resistant mechanisms including Krum \cite{blanchard2017machine}, Bulyan \cite{guerraoui2018hidden}, trimmed mean and median \cite{yin2018byzantine}, which nevertheless focus more on defending untargeted attacks.

\textbf{Visual Contrastive Learning}. Contrastive learning, applied in computer vision tasks, is first proposed to handle unsupervised visual embedding learning problems. Moco \cite{he2020momentum} tries to encode images into key vectors and query vectors. By increasing the similarity between every query vector and its corresponding key vector and decreasing the similarity with other key vectors during training, visual representations with valuable information can be learned. SimCLR \cite{chen2020simple} adopts contrastive learning in self-supervised learning by using data augmentation. SupCon \cite{khosla2020supervised} further utilizes contrastive learning in the supervised setting, which can effectively leverage label information, to improve the model prediction accuracy.

\vspace{-2mm}

\section{What affects backdoor durability in FL?}
\label{method}
In this section, we investigate what affects the durability of backdoors in FL from the perspective of image samples. Instead of trying to explain the vanishing of the backdoors through analyzing interactions between poisoned and benign models, we focus on the impacts of the different relationships among images. In the following part, we define \textit{peer images} as samples who have particular relationships with poisoned images. Further, we identify two types of \textit{peer images} as \textit{interferers} which share the same label with the original label of the poisoned images, and \textit{facilitators} which share the same label with the target label. For example, consider an attacker whose goal is to make the FL model misclassify any green car image as a dog. In this scenario, all car images except for green car images are interferers and all dog images are facilitators. Next, we proceed to reveal the influence of peer images on the durability of backdoors, through carefully designed experiments.
% will reveal the importance of considering the connection between poisoned images and other images through carefully designed experiments. %Also, we will show that two types of \textit{peer images} has completely opposite effect on the fade of the backdoors, and training images except for peer images do not contribute to the durability of backdoors.

\begin{figure}[ht]
\vspace{-2mm}
\setlength{\abovecaptionskip}{0.1in}
\setlength{\belowcaptionskip}{0.1in}
\begin{center}
\centerline{\includegraphics[width=\columnwidth]{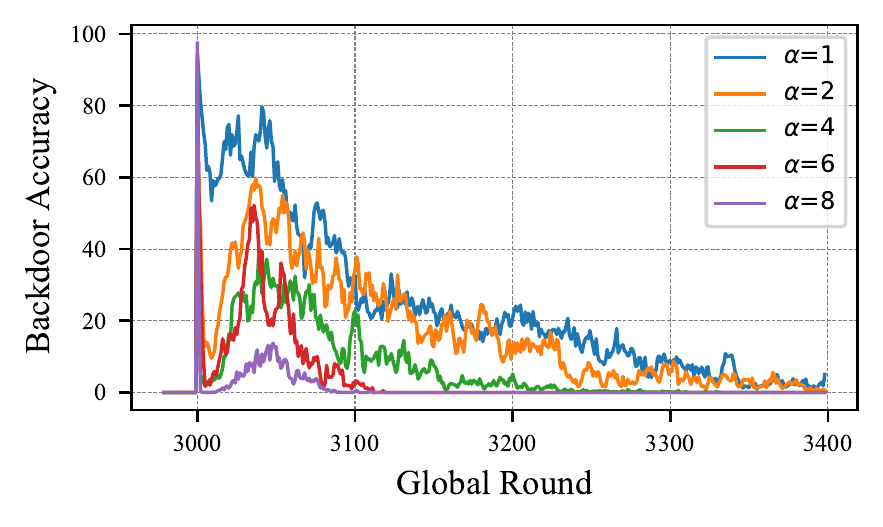}}
\vspace{-5mm}
\caption{Backdoor accuracy with different percentages ($\alpha N/B$) of poisoned images in every training batch. The attacker performs model replacement attack at 3000th global round.}
\label{figure1}
\end{center}
\vspace{-10mm}
\end{figure}

We consider an FL system running an image classification task on CIFAR10 \cite{krizhevsky2009learning}, which is attacked by an extremely strong attacker who can 
%bypass all the defense and successfully perform model replacement attack \cite{bagdasaryan01} and can 
perfectly replace the global model with a poisoned model through model replacement attack \cite{bagdasaryan01}. The poisoned model is trained following \cite{bagdasaryan01}, through creating a malicious training dataset by mixing poisoned images with benign images and perform local training as FedAVG.
%with the original global model. 
The attacker's goal is to make the system misclassify any car-with-vertically-striped-walls-in-the-background image as a bird. A total of $N(N=7)$ poisoned images can be used for training. The attacker trains the poisoned model using mini-batch stochastic gradient descent with batch size $B(B=64)$. The attacker mixes up $\alpha N$ poisoned images and other benign images in every training batch. The variation of backdoor accuracy with global round for different $\alpha$ is shown in \cref{figure1}. The increase in the percentage of poisoned images both lowers the average accuracy and shortens the lifespan of the poison task. Training the poisoned model using only
%the dataset consisting of only 
poisoned images make the model overfit, and the planted backdoor is easier to be erased by any other images. Without involving interferers in the backdoor training, the poisoned model does not pick up the difference between poisoned images and interferers as these two types of images have similar visual features. Thus, the following uploaded benign gradients from interferers are much more likely to conflict with the poisoned model, and then the inserted backdoor fades quickly.

\begin{figure}[hb!]
\vskip -0.2in
\begin{center}
\centerline{\includegraphics[width=\columnwidth]{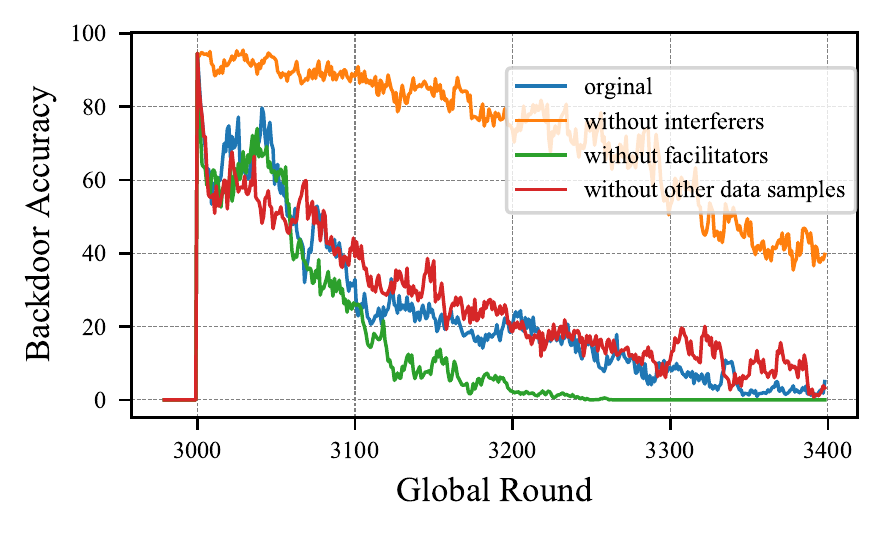}}
\vspace{-5mm}
\caption{Backdoor accuracy with different dataset settings of the following uploaded benign models. %Apart from interferers and facilitators, the absence of any other images in the training dataset of the benign clients does not influence the persistence of the backdoor task. 
The attacker performs model replacement attack at 3000th global round.}
\label{figure2}
\end{center}
%\vskip -0.1in
\vspace{-8mm}
%\vskip -0.2in
\end{figure}

%\vskip -0.1in

We consider another scenario that has the same basic setting as the previous one. Instead, we fix the training dataset of the malicious client by setting $\alpha=1$ and change the training datasets of the following selected benign clients. We deliberately remove certain types of data from the datasets of benign clients to test their effects on backdoor accuracy.
%after the poison round when the global model is replaced with the poisoned model. 
As is shown in \cref{figure2}, 
%we consider four different benign dataset settings and \cref{figure2} indicates
the persistence of the poison task can be significantly enhanced if there are no interferers in the dataset of the following benign clients. However, the lack of facilitators in the dataset of the following benign clients will lead to a decrease in the backdoor durability. Apart from these two types of data, the absence of any other data has minimum influence on the persistence of the backdoor task.

% These two observations provide a basic conception of why it is important to consider the relationship between poisoned images and other images. 
These experimental results illustrate how backdoor durability is affected by interactions between poisoned images with its peer images. Mixing poisoned images and the benign ones in backdoor training helps to enhance the lifespan of the inserted backdoor.
%Instead of solely increase the percentage of poisoned images samples in the poisoned dataset, anticipating other images can help to enhance the lifespan of the inserted backdoor. It also indicates that 
Also, interferers and facilitators play crucial but completely different roles on influencing backdoor durability. We further elaborate in the following.
% Next, we will zoom in to interferers and faciliators to reveal two key factors which affects the durability of backdoors in FL.

\textbf{Update conflict with interferers}. Let the encoder denotes the backbone network of the model. Also, let $T=t_1$ be the global round at which the attacker begins poisoning and $T=t_2$ be the global round in which the attacker finishes lpoisoning. For every poisoned image, it shares similar visual features with interferers (the green car and normal car for example). Without explicit regularization which can guide the encoder to discriminate poisoned data samples from interferers, the encoder may give analogous embedding for them. With analogous embedding and different assigned labels, the updated gradients of poisoned samples and interferers are in conflict. In such case, interferers will inevitably damage poison task performance when $T\in [t_1,t_2]$ and quickly erase the backdoor when $T>t_2$.

\textbf{Slowdown of catastrophic forgetting by facilitators}. Catastrophic forgetting 
%is firstly proposed to illustrate 
refers to the defect of neural networks in learning tasks in a sequential fashion \cite{james01}. Neural networks are inclined to forget previous tasks once the data of these tasks is no longer provided. Similarly, catastrophic forgetting also damages the durability of poison tasks since poisoned images are no longer provided once the attacker leaves the FL system. In the long term, no matter how strong the attacker is, its embedded backdoor will inevitably be erased by the effect of catastrophic forgetting. However, this forgetting effect can be alleviated by facilitators possessed by benign clients. Facilitators, which have the same label as the backdoor target label, in the datasets of following participating benign clients will reintroduce the backdoor information to the global model and extend the lifespan of the inserted backdoor for $T>t_2$.

Motivated by the above observations, the durability of backdoors can be enhanced by leveraging sample relationships between poisoned images and peer images. Specifically, including interferers in the malicious training and making the poisoned images more distinct from the interferers helps the backdoor to last longer. Also, the forgetting effect on backdoors can be further slowed down by making poisoned images more similar to the facilitators. We next propose a novel backdoor attack, Chameleon, to achieve these goals by carefully designing the poisoned dataset, and utilizing contrastive learning in local training to adjust the distances between poisoned images and peer images in favor of a more durable backdoor. 

\begin{algorithm}[h]
   \caption{Chameleon}
   \label{alg1}
\begin{algorithmic}
   \STATE {\bfseries Input:} benign dataset: $D_b$, target label: $y_t$, backdoor type: $t_{bk}$, learning rate in training stage 1 and stage 2: $\eta_1,\ \eta_2$, local batch size: $B$, number of local training rounds in stage 1 and stage 2: $R_1,\ R_2$, local model parameters: 
   %on certain global round: 
   $\theta = (\theta_{enc}, \theta_{cla})$, where $\theta_{enc}$ and $\theta_{cla}$ are the parameters of the encoder and the classifier respectively. 

   \;
   \STATE {\bfseries \itshape Data Preparation Stage}:
   \STATE poisoned data $\boldsymbol{x}_p = \text{CreatePoisonedImg}(t_{bk})$ 
   \STATE poisoned label $y_p = y_t$
   \STATE poisoned dataset $D_p\leftarrow[(\boldsymbol{x}_p,y_p)]$
   \STATE malicious training dataset $\boldsymbol{D}\leftarrow \text{Mix}(D_b,D_p)$

   \;
   \STATE {\bfseries Adaptation Stage:} 
   \FOR{local training round $r=1,\ldots,R_1$}
   \STATE $\mathcal{B}_1\leftarrow$ (split $\boldsymbol{D}$ into batches of size $B$)
   \FOR{$b_1$ $\in$ $\mathcal{B}_1$}
   \STATE $\theta_{enc}=\theta_{enc}-\eta_1\nabla\mathcal{L}_{\text{SupCon}}(\theta_{enc};b_1)$
   \ENDFOR
   \ENDFOR

    \;
   \STATE {\bfseries Projection Stage:} 
   \STATE $\theta_{enc} \leftarrow \text{FreezeParams}(\theta_{enc})$
   \FOR{local training round $r=1,\ldots,R_2$}
   \STATE $\mathcal{B}_2\leftarrow$ (split $\boldsymbol{D}$ into batches of size $B$)
   \FOR{$b_2$ $\in$ $\mathcal{B}_2$}
   \STATE $\theta_{cla}=\theta_{cla}-\eta_2\nabla\mathcal{L}_{\text{CE}}(\theta_{cla};\theta_{enc},b_2)$
   \ENDFOR
   \ENDFOR
   \STATE return $\theta = (\theta_{enc},\theta_{cla})$ to server
\end{algorithmic}
%vspace{-5mm}
\end{algorithm}

\vspace{-3mm}
\section{Enhancing Backdoor Durability Utilizing Sample Relationships}
\subsection{Threat Model}
\label{subsection4_1}
\textbf{Attacker's Capability.} We consider a FL system running image classification tasks, and assume that the attacker can attack the system in a continuous fashion for AttackNum global rounds. In each round, the attacker can corrupt one client. Once corrupted, the attacker has the full control over the training and model uploading process.   We also assume that the system is equipped with norm-clipping defense to eliminate anomalous model updates with irregularly large norms. 

\textbf{Attacker's Goal.} The attacker aims to plant different types of visual backdoors, including pixel-pattern backdoors and semantic backdoors \cite{bagdasaryan01}, into the FL model to make the model misclassify on a certain type of data while leaving other tasks uninfluenced. Further, the attacker would like the embedded backdoor persist as long as possible. Following \cite{Zhengming01}, we use Lifespan to measure the durability of the embedded backdoor.

\textbf{Definition 3.1} (Lifespan). Let $t_0$ be the index of the global round on which the attacker finishes poisoning. Let $\gamma$ be the backdoor threshold accuracy, $\theta_G^t$ be the global model parameters on global round $t$, $f(\cdot)$ be the accuracy function, and let $D_{\text{test}}$ be the test set that contains poisoned images. We can define the $\gamma$-Lifespan $L(\gamma)$ of the backdoor as:
%\vspace{-7mm}
\vskip -0.1in
\begin{equation}
    L(\gamma)=\max \{t|f(\theta_G^t,D_{\text{test}})>\gamma\} - t_0
\end{equation}
%\vskip -0.2in
\vspace{-6mm}
\subsection{Chameleon}
We now introduce our new visual backdoor attack method, Chameleon, which improves backdoors' durability by utilizing the relationships between poisoned images and peer images. In our method, the attacker first prepares a poisoned dataset that mixes benign images and poisoned images. Then, the training on the malicious client proceeds in an adaptation stage and a projection stage. In the adaptation stage, we utilize the contrastive learning paradigm to adjust embeddings of poisoned images and peer images to improve backdoor durability. The projection stage further trains a classifier for specific tasks after freezing the parameters of the encoder learnt from the previous stage. Notably, there is no architecture change between the poisoned model and the global model. \cref{alg1} describes Chameleon in details. %and \cref{figure3} gives graphic illustration of Chameleon. 

% In the following part, we will describe the data preparation stage and two training stages in detail. 
We define $E(\cdot)$ to be the representation encoder which maps input sample $\boldsymbol{x}$ to a embedding vector $\boldsymbol{z} \in \mathbb{R}^d$, and $F(\cdot) = (E(\cdot), C(\cdot))$ to be the deep learning network trained at the clients that starts with $E(\cdot)$ and ends with a classifier $C(\cdot)$.

\textbf{Data Preparation Stage}. %To train the poisoned model, the attacker needs to construct the poisoned set first. 
Let $D_b,\ D_p$ be the benign and poisoned datasets, and $(\boldsymbol{x}_b,y_b),\ (\boldsymbol{x}_p,y_p)$ be the benign and poisoned data-label pairs respectively. Poisoned data $\boldsymbol{x}_p$ is constructed according to different types of backdoors (pixel-pattern and semantic), and poisoned label $y_p$ is the target label. The malicious training dataset $\boldsymbol{D}$ of size $N$ is then constructed by mixing up samples from both the benign dataset $D_b$ and the poisoned dataset $D_p$. Exact mix rates are decided according to specific tasks.

\textbf{Adaptation Stage}. In the first training stage, the corrupted client trains the representation encoder $E(\cdot)$ using supervised contrastive learning \cite{khosla2020supervised}. The corresponding loss function is defined as:

\vspace{-5mm}
\begin{align}
    \mathcal{L}_{\text{SupCon}} \!\!=\! \sum_{i\in I}-\frac{1}{|S(i)|}\!\!\sum_{s\in S(i)} \!\!\log \frac{\beta(i)\exp(\boldsymbol{z}_i\cdot \boldsymbol{z}_s/\tau)}{\sum_{a \in I\backslash \{i\}} \exp(\boldsymbol{z}_i\cdot \boldsymbol{z}_a/\tau)}, \label{con:equ_con}
    %\mathcal{L}_{\text{SupCon}} \!\! 
    %=\! \sum_{i\in I_p\cup I_{Fac}}-\frac{1}{|I_p\cup I_{Fac}|}\!\!\sum_{s\in I_p\cup I_{Fac}} \!\!\log \frac{\beta\exp(\boldsymbol{z}_i\cdot \boldsymbol{z}_s/\tau)}{\sum_{a \in I\backslash \{i\}} \exp(\boldsymbol{z}_i\cdot \boldsymbol{z}_a/\tau)}
    %+\! \sum_{i\in I\backslash \{I_p\cup I_{Fac}\}}-\frac{1}{|S(i)|}\!\!\sum_{s\in S(i)} \!\!\log \frac{\exp(\boldsymbol{z}_i\cdot \boldsymbol{z}_s/\tau)}{\sum_{a \in I\backslash \{i\}} \exp(\boldsymbol{z}_i\cdot \boldsymbol{z}_a/\tau)}\nonumber
\end{align}
\vspace{-4mm}

where $I = \{1,\ldots,N\}$ is the index set of the malicious training dataset $\boldsymbol{D}$, and $S(i)$ is the set of the indices of all samples which share the same label with sample $i$. Also, $\boldsymbol{z_i}$ is the normalized embedding computed by $E(\boldsymbol{x}_i)$ for the input sample $\boldsymbol{x}_i$, and $\beta(i)$ is the weight specifically designed for guiding the encoder to focus more on learning similar embeddings for poisoned images and facilitators. Explicit value of $\beta(i)$ for sample $i$ is defined as:
\vspace{-2.45mm}
\begin{align}
    \beta(i) =
    \begin{cases}
      \beta & y_i = y_p \\
    1 & y_i \neq y_p
    \end{cases},
\end{align}
\vspace{-2.45mm}

where $y_i$ is the label of sample $i$, $y_p$ is the poisoned label, and $\beta >1$ is some design parameter.

\vspace{-3.5mm}

\begin{figure}[ht!]
\begin{center}
\centerline{\includegraphics[width=\columnwidth]{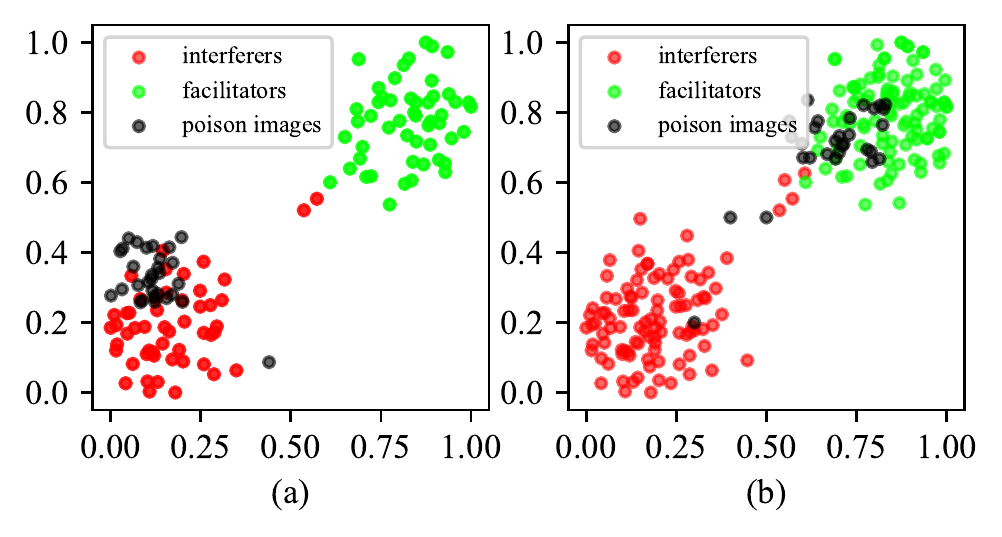}}
\vspace{-5mm}
\caption{Illustration of the 2D embeddings, which are generated using t-SNE \cite{van2008visualizing}, of poisoned images, interferers and facilitators obtained by (a) running the local training of FedAVG and (b) Chameleon.}
\label{figure3}
\end{center}
%\vskip -0.1in
\vspace{-9mm}
\end{figure}

Let $I_p,\ I_{\text{Fac}},\ I_{\text{Int}}$ be the index set of poisoned samples, facilitator samples and interferer samples respectively. For poisoned sample $i\in I_p$, facilitator sample $j\in I_{\text{Fac}}$ and interferer sample $k \in I_{\text{Int}}$, the dot product $\boldsymbol{z}_i\cdot \boldsymbol{z}_j,\ \boldsymbol{z}_i\cdot \boldsymbol{z}_k$ of the normalized embeddings of sample $i,\ j,\ k$ can be seen as the correlation measure of the embeddings between poisoned images and peer images. Optimizing the encoder using the supervised contrastive loss function can drive $\boldsymbol{z}_i\cdot \boldsymbol{z}_j$ to become larger and $\boldsymbol{z}_i\cdot \boldsymbol{z}_k$ to become smaller. Thus, the distance of learned embeddings between poisoned samples and interferers are pushed farther and that between poisoned samples and facilitators are pulled closer. 
As illustrated in an example in \cref{figure3}, compared with normal FedAVG that generates similar embeddings for poisoned images and their interferers (Figure~\ref{figure3}(a)), the adaptation stage of Chameleon moves the poisoned images away from interferers and closer to facilitators (Figure~\ref{figure3}(b)).
% We notice that the learned embeddings of poisoned images are more similar with that of interferers after running the local malicious training of FedAVG in \cref{figure3}(a) while the embeddings of poisoned images and facilitators get more similar after running Chameleon as it is shown in \cref{figure3}(b). 
This helps to both alleviate the update conflict with interferers and strengthen the reintroduction effect of facilitators to compensate for catastrophic forgetting, thus expanding the lifespan of backdoors.

\textbf{Projection Stage}. In the second training stage, the attacker freezes the parameters of the encoder $E(\cdot)$ obtained from the previous stage, and trains a classifier $C(\cdot)$ using the cross-entropy loss $\mathcal{L}_{\text{CE}}$ over $\boldsymbol{D}$ 
%on the top of the representation encoder 
for classification tasks.

\section{Experiments}
\label{exp}
In this section, we provide an extensive empirical evaluation of three different computer vision datasets, CIFAR10, CIFAR100 \cite{krizhevsky2009learning} and EMNIST \cite{cohen2017emnist}, by planting both pixel-pattern and semantic image backdoors \cite{bagdasaryan01} with different FL settings. Our code is available at \url{https://github.com/ybdai7/Chameleon-durable-backdoor}.

We compare with state-of-the-art FL backdoor attacks, including baseline \cite{bagdasaryan01}, Neurotoxin \cite{Zhengming01}, and Anticipate \cite{Yuxin01}:
\vspace{-3mm}
\begin{itemize}[leftmargin=*]
    \item \textbf{Baseline} simply creates the malicious dataset by mixing poisoned images with benign images, and performs local training using projected gradient descent \cite{Hongyi01}.
    \vspace{-2mm}
    \item \textbf{Neurotoxin} tries to improve the durability of the inserted backdoors by identifying the parameters which are not frequently updated by benign clients and inserting backdoors using these parameters.
    \vspace{-2mm}
    \item \textbf{Anticipate} tries to improve the durability of backdoors by considering future model updates from benign clients in the training process of the corrupted client.
\end{itemize}

%Through experiments, we will show that Chameleon outperforms these three methods on durability under all tasks and scenarios.

\subsection{Experiment Setup}
% We implement all tasks in PyTorch \cite{paszke2019pytorch} 
Our experiment implements all the tasks in a FL system running image classification task using FedAVG \cite{mcmahan2017communication}, on a single machine using an NVIDIA GeForce RTX 3090 GPU with 24GB memory. The implemented FL system is assumed to be equipped with norm-clipping defense and the clipping bound $\rho$ is chosen as the maximal value that does not damage the model's convergence performance. %All experiments are conducted on a single machine using an NVIDIA GeForce RTX 3090 GPU with 24GB memory.

Ten randomly selected clients contribute at every aggregation round while the total number of clients is 100. For all three datasets, we randomly split the dataset over clients in a non-iid manner (typical in FL), using Dirichlet sampling with the parameter $\alpha$ set to 0.9 by default \cite{hsu2019measuring}. The durability of backdoor attack methods on certain datasets is evaluated using the $\gamma$-Lifespan $L(\gamma)$ under certain poison accuracy threshold $\gamma$. %Details of each task will be described according to backdoor type and further system setting as follow:

\begin{figure}[ht]
%\vskip 0.1in
\setlength{\abovecaptionskip}{0.1in}
\setlength{\belowcaptionskip}{0.1in}
\begin{center}
\centerline{\includegraphics[width=\columnwidth]{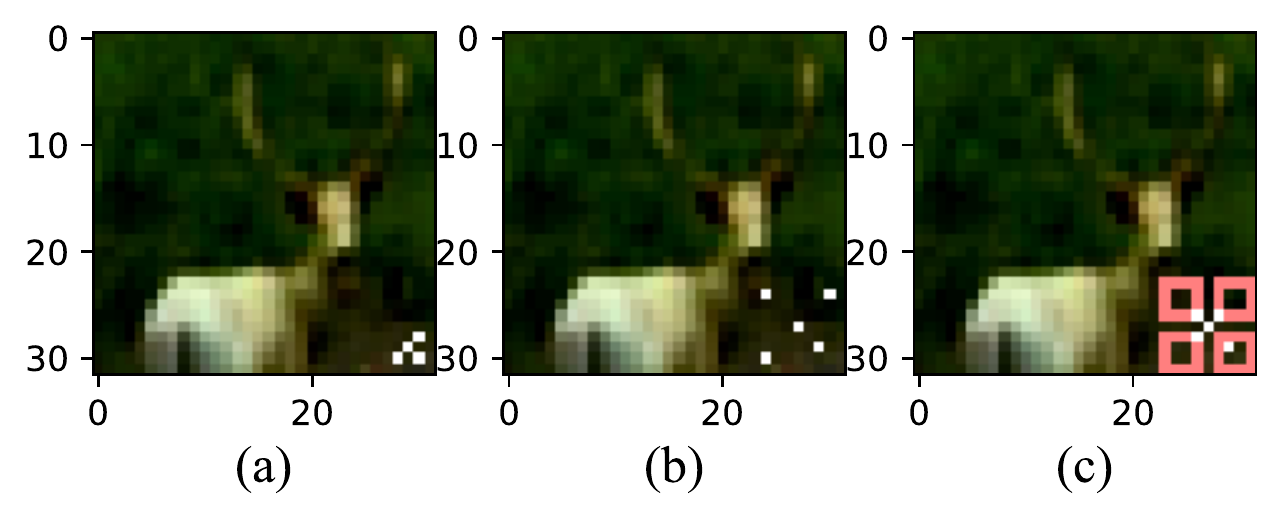}}
\vspace{-5mm}
\caption{(a) Type-1 pixel-pattern backdoor. (b) Type-2 pixel-pattern backdoor. (c) Illustration of possible pixel locations of type-2 pixel-pattern backdoor.}
\label{figure4}
\end{center}
\vspace{-8mm}
\end{figure}

\textbf{Pixel-pattern backdoors}. In this work, we investigate two different types of pixel-pattern backdoors. We choose the first type from \cite{gu2019badnets}, where as shown in \cref{figure4}(a), a fixed pixel-pattern is overlaid with the original image. We also consider a second type to be a dynamic one with the central pixel fixed and locations of the four pixels in the corners are randomly generated within a certain range. One such example in given in \cref{figure4}(b). In general, as illustrated in \cref{figure4}(c), four pixels in the corners can only be generated within its own red box. %which keeps the generated patterns consistent as a whole. 
The coordinates of the corner pixel $(x_e, y_e)$ is generated as follow:
\begin{align} \label{eq_corner_coord}
    (x_e,y_e)= &(x_c \pm (r_c\cdot\text{Random}(0,p_d)+1), \notag \\ 
    & y_c \pm (r_c\cdot\text{Random}(0,p_d)+1)),
\end{align}
where $(x_c,y_c)$ is the coordinate of the central pixel, $\text{Random}(a,b)$ returns a random number between $a$ and $b$, $r_c$ is the maximal range of the corner pixel's coordinate, and $p_d$ represents the pattern diffusion level. With higher $p_d$, the inserted pixel patterns across different poisoned images are more diffused. The sign in Equation \ref{eq_corner_coord} is decided according to the specific location of the corner pixel.

For type-1 pixel-pattern backdoors, we randomly choose images from CIFAR10 and EMNIST and overlay type-1 pixel-pattern triggers on the images to form the poisoned dataset. For type-2 pixel-pattern backdoors, images from CIFAR10 are randomly chosen to be poisoned images with two different pattern diffusion levels $p_d=0.1,\ 0.5$. Poisoned images of CIFAR10 are expected to be classified as birds (label 2) and poisoned images of EMNIST are expected to be classified as number 2 (label 2). We use ResNet18 for these tasks.

\begin{figure}[ht]
\vspace{-1mm}
\setlength{\abovecaptionskip}{0.1in}
\setlength{\belowcaptionskip}{0.1in}
\begin{center}
\centerline{\includegraphics[width=\columnwidth]{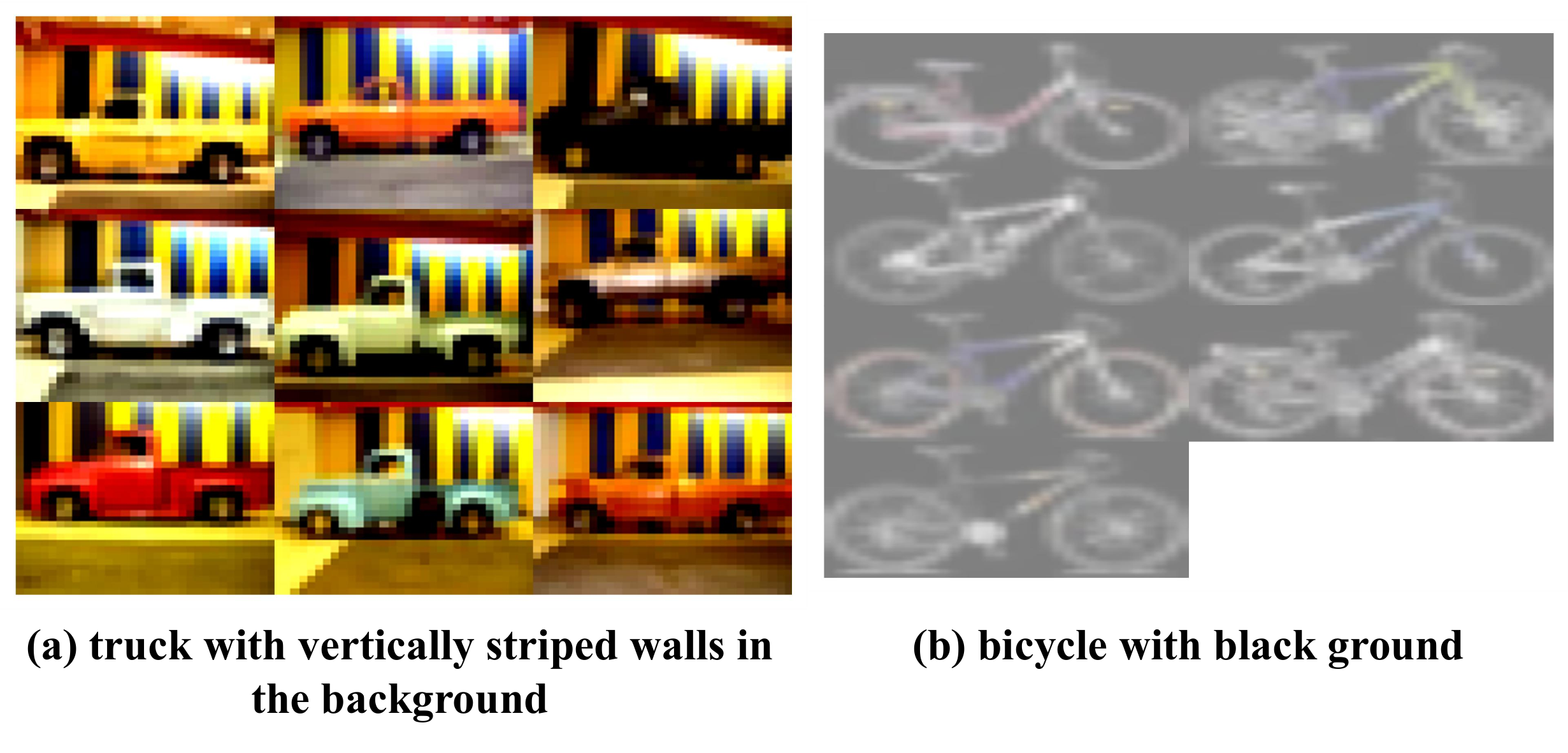}}
\vspace{-3mm}
\caption{Examples of semantic backdoors for CIFAR100 (a) truck-with-vertically-striped-walls-in-the-background. (b) bicycle-with-black-background.}
\label{figure5}
\end{center}
\vspace{-9mm}
\end{figure}

\textbf{Semantic backdoors} can be chosen as any naturally occurring feature of the physical scene and do not require attackers to modify the original images. Following \cite{bagdasaryan01}, we choose green-car, car-with-racing-stripes and car-with-vertically-striped-walls-in-the-background as our semantic backdoors for CIFAR10. All poisoned images in CIFAR10 are expected to be predicted as birds (label 2). Additionally, we choose truck-with-vertically-striped-walls-in-the-background and bicycle-with-black-background as our semantic backdoors for CIFAR100 as shown in \cref{figure5}. Poisoned images with each of the two semantic backdoors are expected to be predicted as butterfly (label 14) and castle (label 17) respectively. We use ResNet18 for these tasks.

\textbf{Model architecture}. We also evaluate the performance of backdoor attacks under different model architectures, including ResNet18 and ResNet34. We use car-with-vertically-striped-walls-in-the-background and car-with-racing-stripes in CIFAR10 as our semantic backdoors.

\subsection{Results}
We present representative results in this section, and leave the others to \cref{appendixA} due to space constraint. We find that in all experiments, Chameleon plants more durable backdoors than all other attacks.
% We will show that Chameleon outperforms prior methods which are baseline, Neurotoxin, and Anticipate, for all tasks under the same level of stealth.

\begin{figure}[h]
%\vskip 0.1in

\begin{center}
\centerline{\includegraphics[width=\columnwidth]{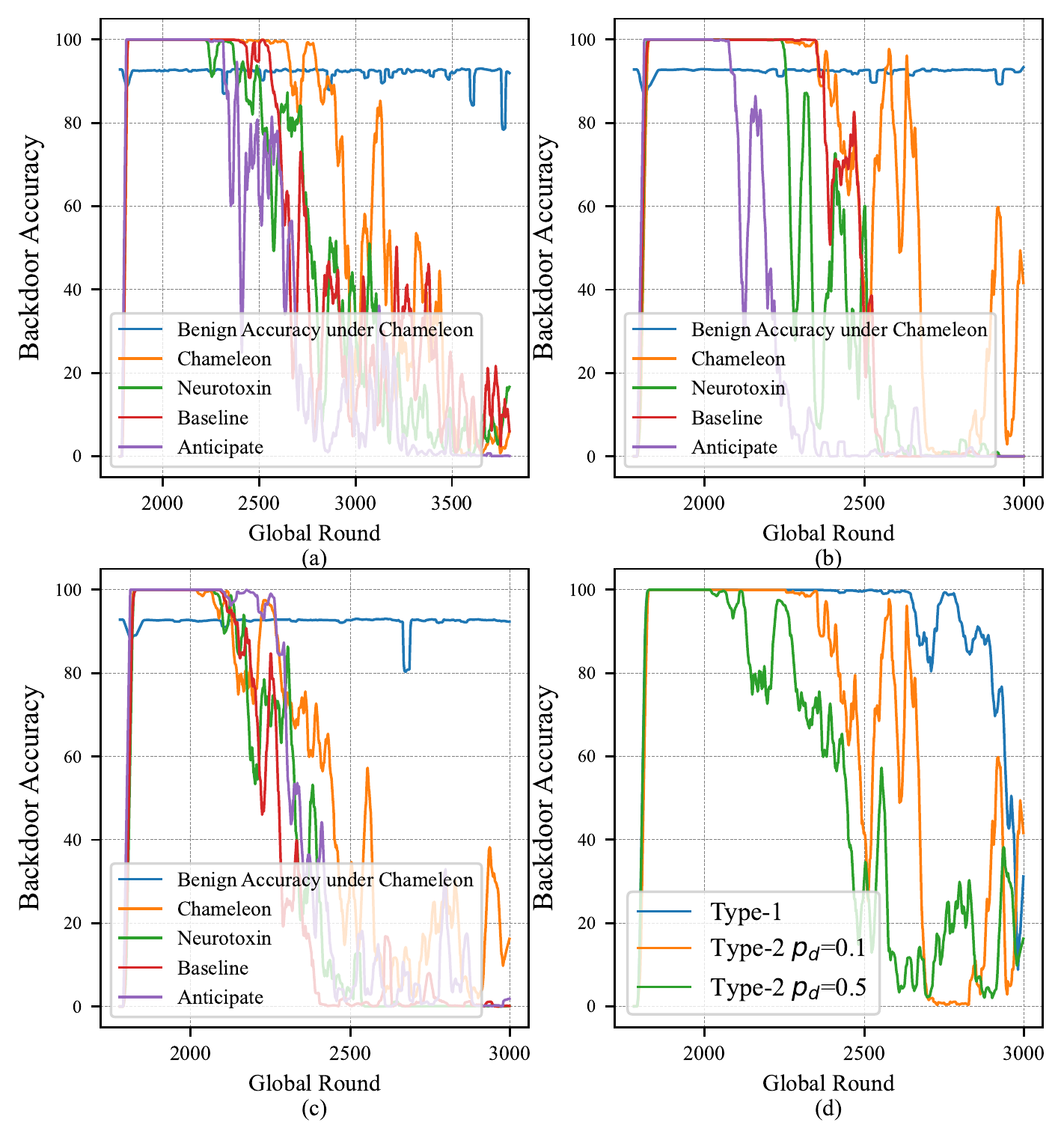}}
\vspace{-5mm}
\caption{Backdoor accuracy on CIFAR10 with (a) Type 1 pixel-pattern backdoor, (b) Type 2 pixel-pattern backdoor with $p_d=0.1$, (c) Type 2 pixel-pattern backdoor with $p_d=0.5$. (d) Chameleon planting pixel-pattern backdoors with different feature diffusion levels. For all tasks, the attack starts at round 1800, and lasts for AttackNum=40 rounds.}
\label{figure6}
\end{center}
\vspace{-8mm}
%\vskip -0.49in
\end{figure}

\textbf{Chameleon improves durability significantly under different settings.} \cref{figure6} presents the results of the performance of different pixel-pattern backdoors on CIFAR10. Chameleon achieves the longest 50\%-Lifespan under all considered pixel-pattern settings in \cref{figure6}(a)-(c). We can also see from \cref{figure6}(d) that type-1 backdoor persists longer than type-2, and the durability of type-2 reduces as $p_d$ increases. 
This indicates that backdoors with more diffused and dynamic features are %usually 
more difficult to plant and are more inclined to vanish. This is because that it becomes harder for the poisoned model to capture backdoor features and learn similar embeddings of poisoned images and facilitators. Thus, the reintroduction effect of facilitators becomes weaker and the lifespan declines.
\vskip -0.1in

\begin{figure}[h!]

\begin{center}
\centerline{\includegraphics[width=\columnwidth]{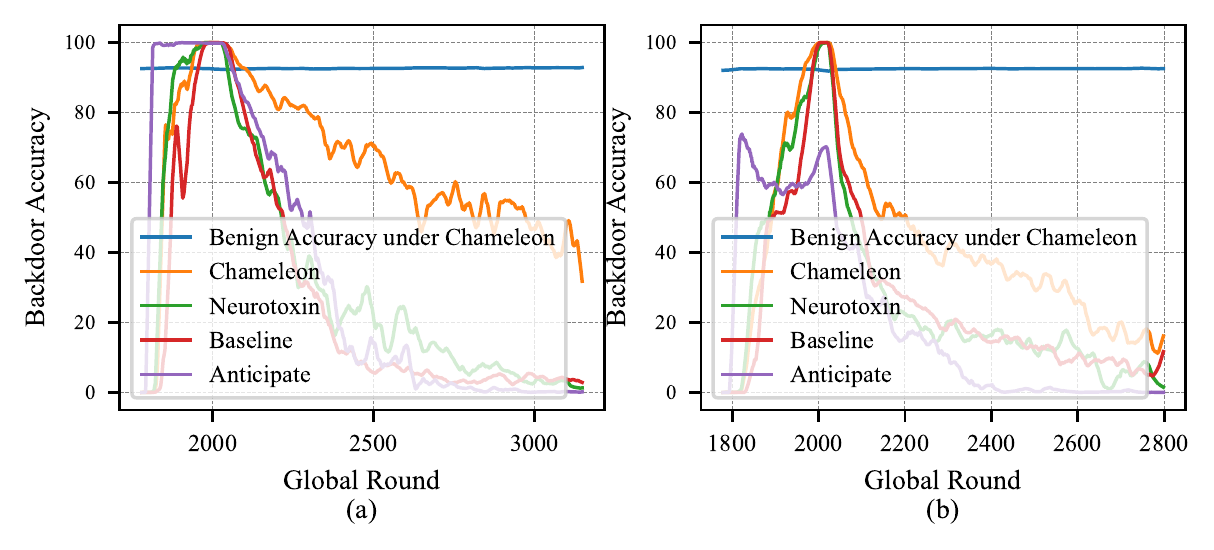}}
\vspace{-4mm}
\caption{Backdoor accuracy on CIFAR10 with (a) car-with-vertically-striped-walls-in-the-background, (b) car-with-racing-stripes. For all tasks, the attack starts at round 1800, and lasts for AttackNum=230 rounds.}
\label{figure7}
\end{center}
\vskip -0.2in
\end{figure}

\vskip -0.2in

\begin{figure}[h!]
%\vskip 0.1in
\begin{center}
\centerline{\includegraphics[width=\columnwidth]{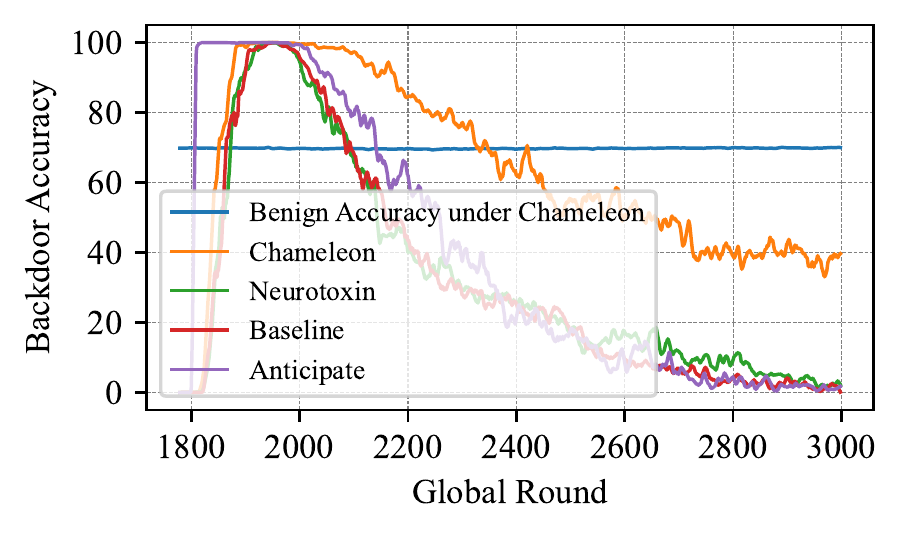}}
\vspace{-4mm}
\caption{Backdoor accuracy on CIFAR100 with bicycle-with-black-background. For this task, the attack starts at round 1800, and lasts for AttackNum=150 rounds.}
\label{figure14}
\end{center}
\vskip -0.3in
\end{figure}

For semantic backdoor experiments, Chameleon improves backdoor durability by $2.8\times\sim4\times$ over prior methods. For two semantic backdoors on CIFAR10, as shown in \cref{figure7}(a) and (b), the 50\%-Lifespans for Chameleon are 1106 and 264, and the max 50\%-Lifespans of the rest three attack methods are 283 and 70, respectively.  We also note that the durability of the car-with-racing-stripes backdoor is weaker than the car-with-vertically-striped-walls-in-the-background backdoor in \cref{figure7}. Similar to the pixel-pattern backdoor, this is because the backdoor features of car-with-racing-stripes images are more diffused (the location and color of the racing stripe are more diverse across poisoned images) than the backdoor features of car-with-vertically-striped-walls-in-the-background images (the location and color of the vertically striped walls are more uniform across poisoned images). %Thus, it is harder for the attacker to plant durable backdoors into the FL system, and 
We note that in \cref{figure7}(b), Anticipate even fails to achieve 100\% backdoor accuracy right after backdoor injection. For CIFAR100 task in \cref{figure14}, the 50\%-Lifespan for Chameleon is 661, improving the max 50\%-Lifespan of the rest three attacks (229) by $2.88 \times$.

\begin{table}[h]
\centering
\caption{50\%-lifespan of the backdoor inserted by different methods (A) Chameleon, (B) Anticipate, (C) Neurotoxin, and (D) Baseline, under different defense methods. Here, - for Anticipate means the method fails to insert backdoor to the FL system under the evaluated defenses. The evaluated backdoor is bicycle-with-black-background samples in CIFAR100.}
\begin{center}
\begin{small}
\begin{tabular}{ccccc}
\toprule  % 顶部线
Defenses&(A)&(B)&(C)&(D) \\ 
\midrule  % 中部线
NCD only&\textbf{661}&229&112&133 \\
Weakly DP \& NCD&\textbf{150}&68&47&- \\
Krum \& NCD&\textbf{72}&-&53&57 \\
FLAME \& NCD&\textbf{102}&-&30&17 \\
\bottomrule  % 底部线
\end{tabular}
\end{small}
\end{center}
\vskip -0.25in
\label{table_new}
\end{table}
    
\textbf{Chameleon improves durability under existing defense methods}. \cref{table_new} demonstrates the effectiveness of our method against three different defense methods: weakly DP \cite{zitengsun01} which is widely adopted in FL defense; Krum \cite{blanchard2017machine} which is the most well-known Byzantine-robust aggregation rule; and FLAME \cite{nguyen2022flame} which is the SOTA backdoor defense method. Under the evaluated defense methods, the durability of the backdoor inserted using Chameleon decreases, compared to the setting where there is only norm-clipping defense (NCD). However, Chameleon still achieves the longest backdoor lifespan among all evaluated methods for the same defense.

\begin{figure}[h]
\vskip -0.1in
\begin{center}
\centerline{\includegraphics[width=\columnwidth]{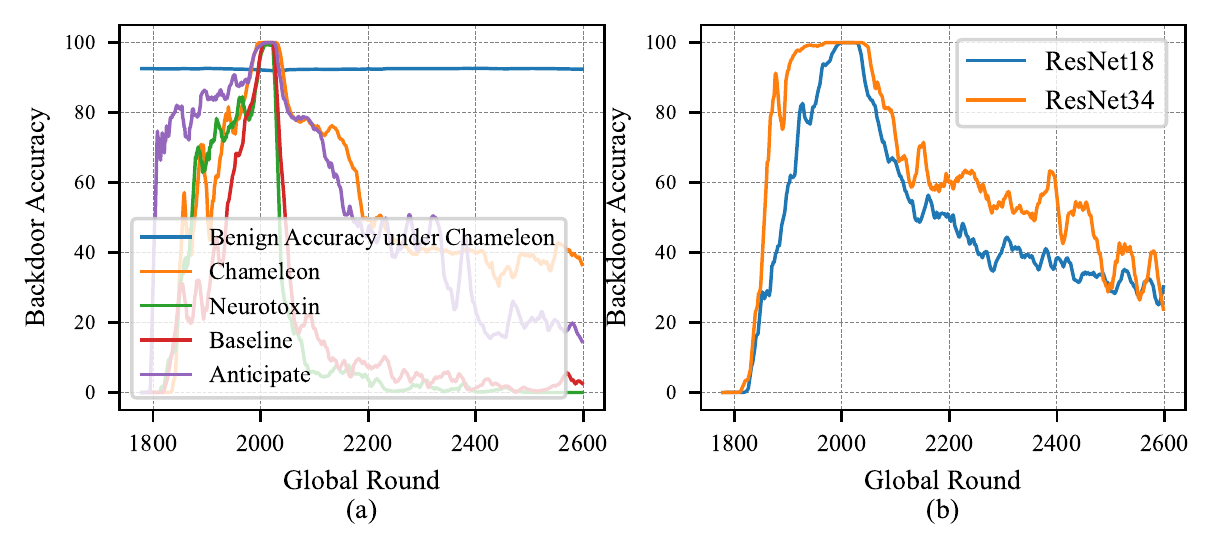}}
\vspace{-5mm}
\caption{(a) Backdoor accuracy using Chameleon on CIFAR10 with car-with-vertically-striped-walls-in-the-background. The model architecture used here is ResNet34. (b) Backdoor accuracy on CIFAR10 with car-with-racing-stripes. The backdoor durability with different model architectures are evaluated. For all tasks, the attack starts at round 1800, and lasts for AttackNum=230 rounds.}
\label{figure8}
\end{center}
\vskip -0.3in
\end{figure}

\textbf{Chameleon is less sensitive to model architectures}. Comparing \cref{figure7}(a) and \cref{figure8}(a), we can see that the model architecture could directly influence the durability of the inserted backdoor. The 40\%-Lifespans for baseline, Neurotoxin reduces from 215, 221 to 13, 7 respectively after the model architecture changes to ResNet34. While the 40\%-Lifespan for Chameleon reduces from 1120 to 567 which is still the longest among all the evaluated attack methods. This demonstrates that compared to prior methods, Chameleon is less sensitive to model architecture and is more suitable for attacking more sophisticated models. 
Interestingly, for backdoors with more diffused features, the durability of backdoors planted by Chameleon improves when the model architecture changes from ResNet18 to ResNet34 in \cref{figure8}(b). We envision that this is because that deeper models possess stronger learning ability to assign distinct embeddings for poisoned images and interferers, and assign similar embeddings for poisoned images and facilitators when the backdoor features get more diffused. %compared to shallow models. In such a case, the attacker can expect this type of backdoors inserted in deeper models to be more durable than shallow models. 

\begin{table}[h]
\vskip -0.2in
\caption{Benign accuracy of the main task on CIFAR10 when Chameleon semantic backdoors (A) car-with-vertically-striped-walls-in-the-background, (B) green-car, and (C) car-with-racing-stripes are planted into the FL system.}
%\label{sample-table}
%\vspace{-3mm}
\vskip -0.2in
\begin{center}
\begin{small}

\begin{tabular}{lcccr}
\toprule
Benign accuracy when & (A) & (B) & (C) \\
\midrule
Start Attack    & 92.55\% & 92.48\% & 92.05\% \\
Stop Attack & 92.25\% & 92.59\% & 92.00\% \\
\bottomrule
\end{tabular}

\end{small}
\label{table1}
\end{center}
\vskip -0.2in
\end{table}

\textbf{Chameleon is stealthy and does not degrade benign accuracy}. \cref{table1} provides the benign accuracy of the main task when the Chameleon attacker plants semantic backdoors into CIFAR10 task. We can see that injecting backdoors do not affect the benign accuracy, which shows that Chameleon is stealthy against human examination on benign accuracy.

\begin{figure}[h!]
\begin{center}
\centerline{\includegraphics[width=\columnwidth]{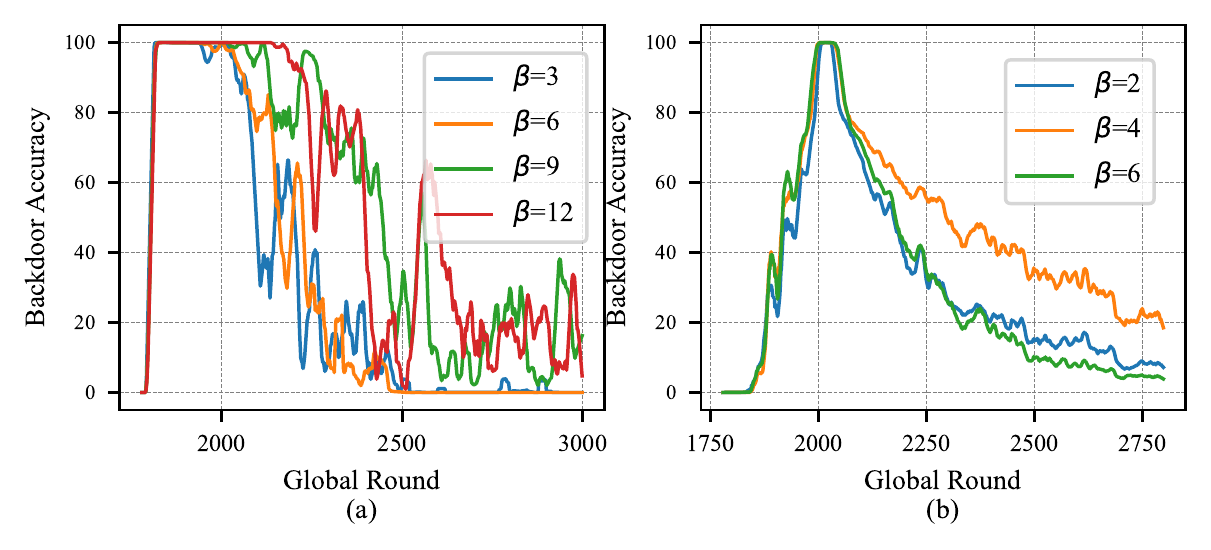}}
\vspace{-4mm}
\caption{The influence of hyper-parameter $\beta$ with different values on the persistence of the inserted (a) type-2 pixel-pattern backdoor with $p_d=0.5$ and (b) semantic backdoor with car-with-racing-stripes trigger. For all tasks, the attack starts at round 1800, and lasts for AttackNum=40, 230 rounds respectively.}
\label{figure13}
\end{center}
\vskip -0.25in
\end{figure}

\begin{table}[h!]
\centering
\vskip -0.2in
\caption{50\%-lifespan of backdoor inserted by different methods under the assumption that data held by different clients in FedAVG is partitioned using Dirichlet sampling with parameter $\alpha$.}
\begin{center}
\begin{small}
\begin{tabular}{ccccc}
\toprule  % 顶部线
$\alpha$&Chameleon&Anticipate&Neurotoxin \\ 
\midrule  % 中部线
0.9 & \textbf{661} & 229 & 112 \\
0.5 & \textbf{261} & 193 & 230 \\
0.2 & \textbf{393} & 137 & 214 \\
\bottomrule  % 底部线
\end{tabular}
\end{small}
\end{center}
\vskip -0.1in
\label{table_new_1}
\end{table}

\vskip -0.1in

\begin{table}[h!]
\vskip -0.1in
\centering
\caption{50\%-lifespan of backdoor inserted by different methods under the assumption that data held by different clients in FedProx is partitioned using Dirichlet sampling with parameter $\alpha$.}
\begin{center}
\begin{small}
\begin{tabular}{ccccc}
\toprule  % 顶部线
$\alpha$&Chameleon&Neurotoxin \\ 
\midrule  % 中部线
0.9 & \textbf{1522} & 472 \\
0.5 & \textbf{422} & 187 \\
0.2 & \textbf{508} & 250 \\
\bottomrule  % 底部线
\end{tabular}
\end{small}
\end{center}
\label{table_new_2}
\vskip -0.1in
\end{table}

\textbf{Chameleon is effective under different data non-iid settings}. When clients' data distribution becomes more non-iid, adversaries may possess fewer interferer and facilitator samples, which may shorten the lifespan of the embedded backdoor. In this case, we can assume that adversaries can manage to collect a small auxiliary dataset $\mathcal{D}_{aux}$ which contains interferer and facilitator samples. Malicious clients can then incorporate a certain number of interferer and facilitator samples drawn from the auxiliary dataset into every training batch to ensure that the embedding of the learnt backdoor is away from that of interferers and close to that of facilitators. %Also please note that other samples except for interferers and facilitators do not contribute to the durability of the embedded backdoor as it is discussed in Section 3. Thus, the durability of the backdoor inserted using our method will not be influenced by different non-iid settings.

The lifespans of multiple backdoor attacks under different degrees of data non-iidness are presented in  \cref{table_new_1} for FedAVG and in \cref{table_new_2} for FedProx  \cite{li2020federated}, which is a popular FL protocol to mitigate model bias due to non-iid data distribution.
We can find that for both FedAVG and FedProx, the durability of the evaluated backdoor, bicycle-with-black-background samples in CIFAR100, inserted by Chameleon is the strongest among all evaluated attacks, for all different values of $\alpha$. Also, for the same value of $\alpha$, Chameleon introduces more durable backdoor in FedProx, compared with FedAVG. 
% the decrease of $\alpha$ generally shortens the backdoor lifespan for the evaluated method. However, in both settings, the durability of the evaluated backdoor, bicycle-with-black-background samples in CIFAR100, inserted by Chameleon is still stronger than the backdoor inserted by Anticipate and Neurotoxin, which demonstrate the effectiveness of Chameleon under different data non-iid settings.

\begin{table}[h!]
\vspace{-5mm}
\centering
\caption{50\%-lifespan of the evaluated backdoors inserted by different methods.}
\begin{center}
\begin{small}
\begin{tabular}{cccc}
\toprule  % 顶部线Baseline
Backdoor&Chameleon&Neurotoxin&Baseline \\ 
\midrule  % 中部线
Blended & \textbf{184} & 60 & 40\\
TCA & \textbf{1566} & 949 & 627\\
\bottomrule  % 底部线
\end{tabular}
\end{small}
\end{center}
\label{table_new_3}
\vspace{-4mm}
%\vskip -0.1in
\end{table}

\textbf{Chameleon is effective to insert more advanced backdoors}. We further provide experimental results on the durability of the blended backdoor \cite{chen2017targeted} and the targeted contamination attack (TCA) \cite{tang2021demon} using different methods. The backdoor trigger chosen for blended backdoor is the randomly generated noise map with mean 0 and variance 0.05. In this scenario, as shown in an example in \cref{figure_new}, \cref{appendixA}, the backdoor images look almost identical to the benign ones. 
%and do not have human-perceptible features. Example image is shown in \cref{figure_new}. 
The backdoor trigger pattern for the targeted contamination backdoor is the same pixel-pattern trigger as that from BadNets \cite{gu2019badnets}. However, instead of being source-agnostic, which means that the poisoned model assigns the target label to samples with backdoor trigger regardless of their original labels, the TCA is source-specific, which only assigns the target class to trigger-carrying samples from certain classes.
From \cref{table_new_3} we can see that the durability of the backdoor inserted by Chameleon is still the strongest among the evaluated methods, for these two advanced backdoors.
%which shows that Chameleon is effective to insert a wide range of backdoors

\textbf{Ablation study on $\beta$}. We investigate the influence of choosing different $\beta$, which is the hyper-parameter in the contrastive learning loss function in (\ref{con:equ_con}), on the durability of Chameleon backdoors. \cref{figure13} shows the influence of different $\beta$ on pixel-pattern backdoor and semantic backdoor. 

As shown in \cref{figure13}(a), the durability of type-2 pixel-pattern backdoor improves monotonically as $\beta$ increase from 3 to 12. Whereas
for semantic backdoor in \cref{figure13}(b), the strongest durability is achieved with an intermediate value of $\beta=4$. This can be explained from the internal difference between pixel-pattern backdoors and semantic backdoors. For pixel-pattern backdoors, the attacker need to overlay carefully chosen triggers on the original images to generated poisoned images. It is thus easier for the poisoned model to learn the difference between poisoned images and interferers, and assigns more distinct embeddings for them. Update conflicts with interferers are then significantly alleviated. In this case, the attacker can improve backdoor durability mainly through strengthening the reintroduction effect of facilitators, and this can be achieved by increasing $\beta$ to further pull closer embeddings of poisoned images and facilitators during contrastive learning. However, update conflicts with interferers are more severe in semantic backdoors as we do not insert new backdoor trigger and consider some naturally occurring feature as the backdoor trigger. This makes assigning distinct embeddings for poisoned images and interferers more difficult, and update conflicts with interferers can greatly damage the durability in this case. Unlike the case of the case of pixel-pattern backdoors, increasing $\beta$ introduces contradictory effects on moving away from interferers and moving towards facilitators. Therefore, to maximize backdoor durability, a value of $\beta$ that strikes a good balance between these two effects should be chosen.
% In sharp contrast to the case of pixel-pattern backdoors, a large $\beta$ will make update conflicts with interferers more severe, leading to the decrease of the lifespan. So it is not optimal to keep increasing $\beta$ for semantic backdoors. Instead, the value of $\beta$ which strikes a good balance between update conflicts with interferers and the reintroduction effect of facilitators should be chosen.

\section{Discussion and Conclusion}
\label{diss}
Notably, the main focus of Chameleon is to extend the lifespan of backdoors in FL but not in general to escape all advanced backdoor defenses. We envision that future researches that aim to escape from advanced defenses can combine with our method to improve the durability of the inserted backdoors.

In this paper, we find that utilizing the sample relationships can greatly enhance the durability of the inserted visual backdoor in FL. Based on the observation, we developed a novel attack method, Chameleon, which adopts contrastive learning paradigms to adjust the relative embedding relationship among poisoned samples, interferers, and facilitators to improve the durability of backdoors. Over evaluations for a wide range of datasets, backdoor types and model architectures, we show that Chameleon are consistently durable and can improve the backdoor durability by $1.2\times\sim4\times$ compared to prior methods.

\vspace{-3mm}
\section*{Acknowledgement}
\vspace{-2mm}
This work is in part supported by the National Nature Science Foundation of China (NSFC) Grant 62106057, Guangzhou Municipal Science and Technology Guangzhou-HKUST(GZ) Joint Project 2023A03J0151 and Project 2023A03J0011, Foshan HKUST Projects FSUST20-FYTRI04B, and Guangdong Provincial Key Lab of Integrated Communication, Sensing and Computation for Ubiquitous Internet of Things.

\bibliography{chameleon}

\begin{thebibliography}{43}
\providecommand{\natexlab}[1]{#1}
\providecommand{\url}[1]{\texttt{#1}}
\expandafter\ifx\csname urlstyle\endcsname\relax
  \providecommand{\doi}[1]{doi: #1}\else
  \providecommand{\doi}{doi: \begingroup \urlstyle{rm}\Url}\fi

\bibitem[Andreina et~al.(2021)Andreina, Marson, M{\"o}llering, and
  Karame]{andreina2021baffle}
Andreina, S., Marson, G.~A., M{\"o}llering, H., and Karame, G.
\newblock Baffle: Backdoor detection via feedback-based federated learning.
\newblock In \emph{2021 IEEE 41st International Conference on Distributed
  Computing Systems (ICDCS)}, pp.\  852--863. IEEE, 2021.

\bibitem[Bagdasaryan et~al.(2020)Bagdasaryan, Veit, Hua, Estrin, and
  Shmatikov]{bagdasaryan01}
Bagdasaryan, E., Veit, A., Hua, Y., Estrin, D., and Shmatikov, V.
\newblock How to backdoor federated learning.
\newblock In \emph{International Conference on Artificial Intelligence and
  Statistics}, pp.\  2938--2948. PMLR, 2020.

\bibitem[Bhagoji et~al.(2019)Bhagoji, Chakraborty, Mittal, and
  Calo]{bhagoji2019analyzing}
Bhagoji, A.~N., Chakraborty, S., Mittal, P., and Calo, S.
\newblock Analyzing federated learning through an adversarial lens.
\newblock In \emph{International Conference on Machine Learning}, pp.\
  634--643. PMLR, 2019.

\bibitem[Blanchard et~al.(2017)Blanchard, El~Mhamdi, Guerraoui, and
  Stainer]{blanchard2017machine}
Blanchard, P., El~Mhamdi, E.~M., Guerraoui, R., and Stainer, J.
\newblock Machine learning with adversaries: Byzantine tolerant gradient
  descent.
\newblock \emph{Advances in Neural Information Processing Systems}, 30, 2017.

\bibitem[Chen et~al.(2020)Chen, Kornblith, Norouzi, and Hinton]{chen2020simple}
Chen, T., Kornblith, S., Norouzi, M., and Hinton, G.
\newblock A simple framework for contrastive learning of visual
  representations.
\newblock In \emph{International conference on machine learning}, pp.\
  1597--1607. PMLR, 2020.

\bibitem[Chen et~al.(2017{\natexlab{a}})Chen, Liu, Li, Lu, and
  Song]{chen2017targeted}
Chen, X., Liu, C., Li, B., Lu, K., and Song, D.
\newblock Targeted backdoor attacks on deep learning systems using data
  poisoning.
\newblock \emph{arXiv preprint arXiv:1712.05526}, 2017{\natexlab{a}}.

\bibitem[Chen et~al.(2017{\natexlab{b}})Chen, Su, and Xu]{chen2017distributed}
Chen, Y., Su, L., and Xu, J.
\newblock Distributed statistical machine learning in adversarial settings:
  Byzantine gradient descent.
\newblock \emph{Proceedings of the ACM on Measurement and Analysis of Computing
  Systems}, 1\penalty0 (2):\penalty0 1--25, 2017{\natexlab{b}}.

\bibitem[Cohen et~al.(2017)Cohen, Afshar, Tapson, and
  Van~Schaik]{cohen2017emnist}
Cohen, G., Afshar, S., Tapson, J., and Van~Schaik, A.
\newblock Emnist: Extending mnist to handwritten letters.
\newblock In \emph{2017 international joint conference on neural networks
  (IJCNN)}, pp.\  2921--2926. IEEE, 2017.

\bibitem[Fang et~al.(2020)Fang, Cao, Jia, and Gong]{fang2020local}
Fang, M., Cao, X., Jia, J., and Gong, N.~Z.
\newblock Local model poisoning attacks to byzantine-robust federated learning.
\newblock In \emph{Proceedings of the 29th USENIX Conference on Security
  Symposium}, pp.\  1623--1640, 2020.

\bibitem[Fung et~al.(2020)Fung, Yoon, and Beschastnikh]{fung2020limitations}
Fung, C., Yoon, C.~J., and Beschastnikh, I.
\newblock The limitations of federated learning in sybil settings.
\newblock In \emph{RAID}, pp.\  301--316, 2020.

\bibitem[Geiping et~al.(2020)Geiping, Bauermeister, Dr{\"o}ge, and
  Moeller]{geiping2020inverting}
Geiping, J., Bauermeister, H., Dr{\"o}ge, H., and Moeller, M.
\newblock Inverting gradients-how easy is it to break privacy in federated
  learning?
\newblock \emph{Advances in Neural Information Processing Systems},
  33:\penalty0 16937--16947, 2020.

\bibitem[Gu et~al.(2019)Gu, Liu, Dolan-Gavitt, and Garg]{gu2019badnets}
Gu, T., Liu, K., Dolan-Gavitt, B., and Garg, S.
\newblock Badnets: Evaluating backdooring attacks on deep neural networks.
\newblock \emph{IEEE Access}, 7:\penalty0 47230--47244, 2019.

\bibitem[Guerraoui et~al.(2018)Guerraoui, Rouault, et~al.]{guerraoui2018hidden}
Guerraoui, R., Rouault, S., et~al.
\newblock The hidden vulnerability of distributed learning in byzantium.
\newblock In \emph{International Conference on Machine Learning}, pp.\
  3521--3530. PMLR, 2018.

\bibitem[He et~al.(2020)He, Fan, Wu, Xie, and Girshick]{he2020momentum}
He, K., Fan, H., Wu, Y., Xie, S., and Girshick, R.
\newblock Momentum contrast for unsupervised visual representation learning.
\newblock In \emph{Proceedings of the IEEE/CVF conference on computer vision
  and pattern recognition}, pp.\  9729--9738, 2020.

\bibitem[Hsu et~al.(2019)Hsu, Qi, and Brown]{hsu2019measuring}
Hsu, T.-M.~H., Qi, H., and Brown, M.
\newblock Measuring the effects of non-identical data distribution for
  federated visual classification.
\newblock \emph{arXiv preprint arXiv:1909.06335}, 2019.

\bibitem[Khosla et~al.(2020)Khosla, Teterwak, Wang, Sarna, Tian, Isola,
  Maschinot, Liu, and Krishnan]{khosla2020supervised}
Khosla, P., Teterwak, P., Wang, C., Sarna, A., Tian, Y., Isola, P., Maschinot,
  A., Liu, C., and Krishnan, D.
\newblock Supervised contrastive learning.
\newblock \emph{Advances in Neural Information Processing Systems},
  33:\penalty0 18661--18673, 2020.

\bibitem[Kirkpatrick et~al.(2017)Kirkpatrick, Pascanu, Rabinowitz, Veness,
  Desjardins, Rusu, Milan, Quan, Ramalho, Grabska-Barwinska, et~al.]{james01}
Kirkpatrick, J., Pascanu, R., Rabinowitz, N., Veness, J., Desjardins, G., Rusu,
  A.~A., Milan, K., Quan, J., Ramalho, T., Grabska-Barwinska, A., et~al.
\newblock Overcoming catastrophic forgetting in neural networks.
\newblock \emph{Proceedings of the national academy of sciences}, 114\penalty0
  (13):\penalty0 3521--3526, 2017.

\bibitem[Krizhevsky et~al.(2009)Krizhevsky, Hinton,
  et~al.]{krizhevsky2009learning}
Krizhevsky, A., Hinton, G., et~al.
\newblock Learning multiple layers of features from tiny images.
\newblock 2009.

\bibitem[Li et~al.(2020)Li, Sahu, Zaheer, Sanjabi, Talwalkar, and
  Smith]{li2020federated}
Li, T., Sahu, A.~K., Zaheer, M., Sanjabi, M., Talwalkar, A., and Smith, V.
\newblock Federated optimization in heterogeneous networks.
\newblock \emph{Proceedings of Machine Learning and Systems}, 2:\penalty0
  429--450, 2020.

\bibitem[Liu et~al.(2018)Liu, Ma, Aafer, Lee, Zhai, Wang, and
  Zhang]{liu2017trojaning}
Liu, Y., Ma, S., Aafer, Y., Lee, W.-C., Zhai, J., Wang, W., and Zhang, X.
\newblock Trojaning attack on neural networks.
\newblock In \emph{25th Annual Network And Distributed System Security
  Symposium (NDSS 2018)}. Internet Soc, 2018.

\bibitem[Liu et~al.(2020)Liu, Ma, Bailey, and Lu]{liu2020reflection}
Liu, Y., Ma, X., Bailey, J., and Lu, F.
\newblock Reflection backdoor: A natural backdoor attack on deep neural
  networks.
\newblock In \emph{Computer Vision--ECCV 2020: 16th European Conference,
  Glasgow, UK, August 23--28, 2020, Proceedings, Part X 16}, pp.\  182--199.
  Springer, 2020.

\bibitem[Lyu et~al.(2022)Lyu, Yu, Ma, Chen, Sun, Zhao, Yang, and
  Philip]{lyu2022privacy}
Lyu, L., Yu, H., Ma, X., Chen, C., Sun, L., Zhao, J., Yang, Q., and Philip,
  S.~Y.
\newblock Privacy and robustness in federated learning: Attacks and defenses.
\newblock \emph{IEEE transactions on neural networks and learning systems},
  2022.

\bibitem[McMahan et~al.(2017)McMahan, Moore, Ramage, Hampson, and
  y~Arcas]{mcmahan2017communication}
McMahan, B., Moore, E., Ramage, D., Hampson, S., and y~Arcas, B.~A.
\newblock Communication-efficient learning of deep networks from decentralized
  data.
\newblock In \emph{Artificial intelligence and statistics}, pp.\  1273--1282.
  PMLR, 2017.

\bibitem[Naseri et~al.(2022)Naseri, Hayes, and De~Cristofaro]{naseri2022local}
Naseri, M., Hayes, J., and De~Cristofaro, E.
\newblock Local and central differential privacy for robustness and privacy in
  federated learning.
\newblock In \emph{NDSS}, 2022.

\bibitem[Nasr et~al.(2019)Nasr, Shokri, and Houmansadr]{nasr2019comprehensive}
Nasr, M., Shokri, R., and Houmansadr, A.
\newblock Comprehensive privacy analysis of deep learning: Passive and active
  white-box inference attacks against centralized and federated learning.
\newblock In \emph{2019 IEEE symposium on security and privacy (SP)}, pp.\
  739--753. IEEE, 2019.

\bibitem[Nguyen et~al.(2022)Nguyen, Rieger, Chen, Yalame, M{\"o}llering,
  Fereidooni, Marchal, Miettinen, Mirhoseini, Zeitouni,
  et~al.]{nguyen2022flame}
Nguyen, T.~D., Rieger, P., Chen, H., Yalame, H., M{\"o}llering, H., Fereidooni,
  H., Marchal, S., Miettinen, M., Mirhoseini, A., Zeitouni, S., et~al.
\newblock $\{$FLAME$\}$: Taming backdoors in federated learning.
\newblock In \emph{31st USENIX Security Symposium (USENIX Security 22)}, pp.\
  1415--1432, 2022.

\bibitem[Rieger et~al.(2022{\natexlab{a}})Rieger, Krau{\ss}, Miettinen,
  Dmitrienko, and Sadeghi]{rieger2022close}
Rieger, P., Krau{\ss}, T., Miettinen, M., Dmitrienko, A., and Sadeghi, A.-R.
\newblock Close the gate: Detecting backdoored models in federated learning
  based on client-side deep layer output analysis.
\newblock \emph{arXiv preprint arXiv:2210.07714}, 2022{\natexlab{a}}.

\bibitem[Rieger et~al.(2022{\natexlab{b}})Rieger, Nguyen, Miettinen, and
  Sadeghi]{rieger2022deepsight}
Rieger, P., Nguyen, T.~D., Miettinen, M., and Sadeghi, A.-R.
\newblock Deepsight: Mitigating backdoor attacks in federated learning through
  deep model inspection.
\newblock In \emph{NDSS}, 2022{\natexlab{b}}.

\bibitem[Saha et~al.(2020)Saha, Subramanya, and Pirsiavash]{saha2020hidden}
Saha, A., Subramanya, A., and Pirsiavash, H.
\newblock Hidden trigger backdoor attacks.
\newblock In \emph{Proceedings of the AAAI conference on artificial
  intelligence}, volume~34, pp.\  11957--11965, 2020.

\bibitem[Shafahi et~al.(2018)Shafahi, Huang, Najibi, Suciu, Studer, Dumitras,
  and Goldstein]{shafahi2018poison}
Shafahi, A., Huang, W.~R., Najibi, M., Suciu, O., Studer, C., Dumitras, T., and
  Goldstein, T.
\newblock Poison frogs! targeted clean-label poisoning attacks on neural
  networks.
\newblock \emph{Advances in neural information processing systems}, 31, 2018.

\bibitem[Shejwalkar \& Houmansadr(2021)Shejwalkar and
  Houmansadr]{shejwalkar2021manipulating}
Shejwalkar, V. and Houmansadr, A.
\newblock Manipulating the byzantine: Optimizing model poisoning attacks and
  defenses for federated learning.
\newblock In \emph{NDSS}, 2021.

\bibitem[Shejwalkar et~al.(2022)Shejwalkar, Houmansadr, Kairouz, and
  Ramage]{shejwalkar2022back}
Shejwalkar, V., Houmansadr, A., Kairouz, P., and Ramage, D.
\newblock Back to the drawing board: A critical evaluation of poisoning attacks
  on production federated learning.
\newblock In \emph{2022 IEEE Symposium on Security and Privacy (SP)}, pp.\
  1354--1371. IEEE, 2022.

\bibitem[Shen et~al.(2016)Shen, Tople, and Saxena]{shen2016auror}
Shen, S., Tople, S., and Saxena, P.
\newblock Auror: Defending against poisoning attacks in collaborative deep
  learning systems.
\newblock In \emph{Proceedings of the 32nd Annual Conference on Computer
  Security Applications}, pp.\  508--519, 2016.

\bibitem[Sun et~al.(2019)Sun, Kairouz, Suresh, and McMahan]{zitengsun01}
Sun, Z., Kairouz, P., Suresh, A.~T., and McMahan, H.~B.
\newblock Can you really backdoor federated learning?
\newblock \emph{arXiv preprint arXiv:1911.07963}, 2019.

\bibitem[Tang et~al.(2021)Tang, Wang, Tang, and Zhang]{tang2021demon}
Tang, D., Wang, X., Tang, H., and Zhang, K.
\newblock Demon in the variant: Statistical analysis of dnns for robust
  backdoor contamination detection.
\newblock In \emph{USENIX Security Symposium}, pp.\  1541--1558, 2021.

\bibitem[Van~der Maaten \& Hinton(2008)Van~der Maaten and
  Hinton]{van2008visualizing}
Van~der Maaten, L. and Hinton, G.
\newblock Visualizing data using t-sne.
\newblock \emph{Journal of machine learning research}, 9\penalty0 (11), 2008.

\bibitem[Wang et~al.(2020)Wang, Sreenivasan, Rajput, Vishwakarma, Agarwal,
  Sohn, Lee, and Papailiopoulos]{Hongyi01}
Wang, H., Sreenivasan, K., Rajput, S., Vishwakarma, H., Agarwal, S., Sohn,
  J.-y., Lee, K., and Papailiopoulos, D.
\newblock Attack of the tails: Yes, you really can backdoor federated learning.
\newblock \emph{Advances in Neural Information Processing Systems},
  33:\penalty0 16070--16084, 2020.

\bibitem[Wen et~al.(2022)Wen, Geiping, Fowl, Souri, Chellappa, Goldblum, and
  Goldstein]{Yuxin01}
Wen, Y., Geiping, J., Fowl, L., Souri, H., Chellappa, R., Goldblum, M., and
  Goldstein, T.
\newblock Thinking two moves ahead: Anticipating other users improves backdoor
  attacks in federated learning.
\newblock \emph{arXiv preprint arXiv:2210.09305}, 2022.

\bibitem[Xie et~al.(2019)Xie, Huang, Chen, and Li]{xie2019dba}
Xie, C., Huang, K., Chen, P.-Y., and Li, B.
\newblock Dba: Distributed backdoor attacks against federated learning.
\newblock In \emph{International Conference on Learning Representations}, 2019.

\bibitem[Xie et~al.(2021)Xie, Chen, Chen, and Li]{xie2021crfl}
Xie, C., Chen, M., Chen, P.-Y., and Li, B.
\newblock Crfl: Certifiably robust federated learning against backdoor attacks.
\newblock In \emph{International Conference on Machine Learning}, pp.\
  11372--11382. PMLR, 2021.

\bibitem[Yin et~al.(2018)Yin, Chen, Kannan, and Bartlett]{yin2018byzantine}
Yin, D., Chen, Y., Kannan, R., and Bartlett, P.
\newblock Byzantine-robust distributed learning: Towards optimal statistical
  rates.
\newblock In \emph{International Conference on Machine Learning}, pp.\
  5650--5659. PMLR, 2018.

\bibitem[Zhang et~al.(2022)Zhang, Panda, Song, Yang, Mahoney, Mittal, Kannan,
  and Gonzalez]{Zhengming01}
Zhang, Z., Panda, A., Song, L., Yang, Y., Mahoney, M., Mittal, P., Kannan, R.,
  and Gonzalez, J.
\newblock Neurotoxin: Durable backdoors in federated learning.
\newblock In \emph{International Conference on Machine Learning}, pp.\
  26429--26446. PMLR, 2022.

\bibitem[Zhao et~al.(2020)Zhao, Hu, Wang, Jiang, Shen, Luo, and
  Hu]{zhao2020shielding}
Zhao, L., Hu, S., Wang, Q., Jiang, J., Shen, C., Luo, X., and Hu, P.
\newblock Shielding collaborative learning: Mitigating poisoning attacks
  through client-side detection.
\newblock \emph{IEEE Transactions on Dependable and Secure Computing},
  18\penalty0 (5):\penalty0 2029--2041, 2020.

\end{thebibliography}
\bibliographystyle{icml2023}

%%%%%%%%%%%%%%%%%%%%%%%%%%%%%%%%%%%%%%%%%%%%%%%%%%%%%%%%%%%%%%%%%%%%%%%%%%%%%%%
%%%%%%%%%%%%%%%%%%%%%%%%%%%%%%%%%%%%%%%%%%%%%%%%%%%%%%%%%%%%%%%%%%%%%%%%%%%%%%%
% APPENDIX
%%%%%%%%%%%%%%%%%%%%%%%%%%%%%%%%%%%%%%%%%%%%%%%%%%%%%%%%%%%%%%%%%%%%%%%%%%%%%%%
%%%%%%%%%%%%%%%%%%%%%%%%%%%%%%%%%%%%%%%%%%%%%%%%%%%%%%%%%%%%%%%%%%%%%%%%%%%%%%%
\newpage
\appendix
\onecolumn
\section{Complementary experiment results}
\label{appendixA}
In this section, we will provide experimental results and analysis of all the tasks which are not presented in the main text and summarize the lifespan of all the tasks under different attack methods to further prove the superiority of our proposed method.

\vspace{-2mm}

\begin{figure}[h!]
\begin{center}
\centerline{\includegraphics{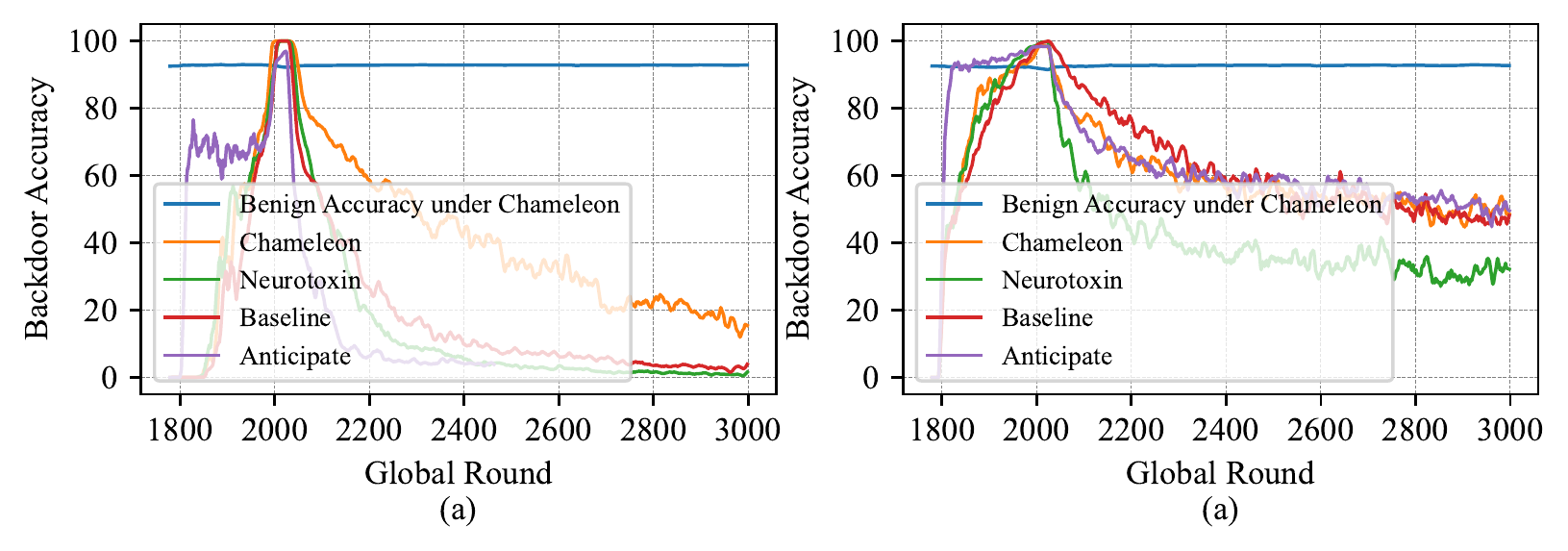}}
\vspace{-3mm}
\caption{Backdoor accuracy on CIFAR10 with (a) green-car, (b) southwest airline images from edge-case dataset. For all tasks, the attack starts at round 1800, and lasts for AttackNum=230 rounds.}
\label{figure9}
\end{center}
\vskip -0.1in
\end{figure}

\vspace{-3mm}

\begin{figure}[h!]
\begin{center}
\centerline{\includegraphics{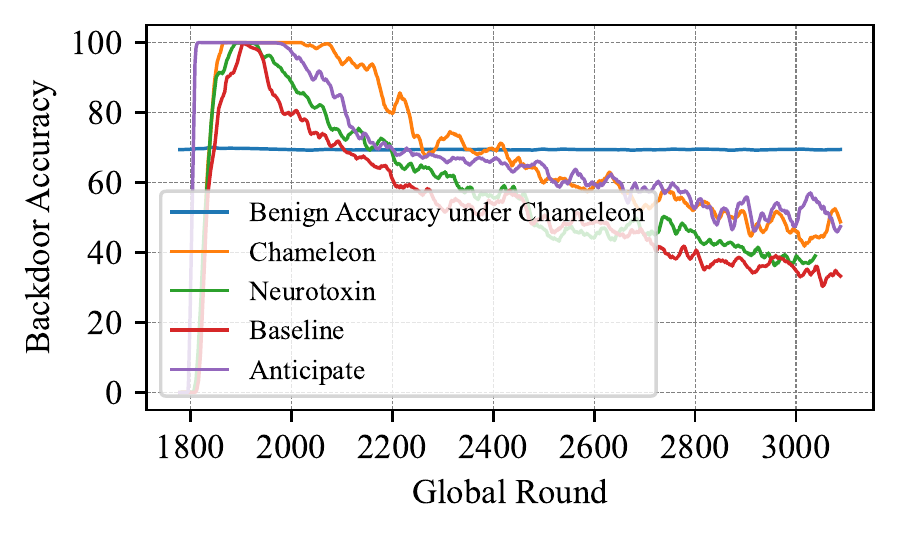}}
\vspace{-3mm}
\caption{Backdoor accuracy on CIFAR100 with truck-with-vertically-striped-walls-in-the-background. For all tasks, the attack starts at round 1800, and lasts for AttackNum=100 rounds.}
\label{figure10}
\end{center}
\vskip -0.1in
\end{figure}

\vspace{-3mm}

\begin{figure}[h!]
\begin{center}
\vspace{-3mm}
\centerline{\includegraphics{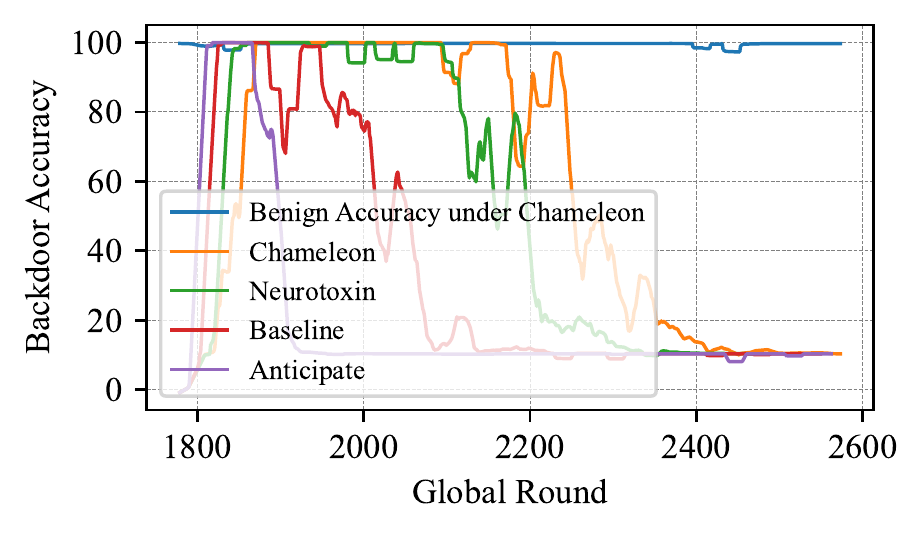}}
\caption{Backdoor accuracy on EMNIST with type-1 pixel-pattern backdoor. For all tasks, the attack starts at round 1800, and lasts for AttackNum=60 rounds.}
\label{figure11}
\end{center}
\vskip -0.2in
\end{figure}

\vspace{-2mm}

\begin{table}[h!]
\caption{Lifespan $L(\gamma)$ of the backdoor inserted by Chameleon, Anticipate, Neurotoxin and baseline of all tasks evaluated in this paper. The model architecture used here is ResNet18.}
\begin{center}
\begin{small}
\begin{tabular}{lcccccr}
\toprule
Datasets & Backdoor & AttckNum & Chameleon & Anticipate & Neurotoxin & Baseline \\
\midrule
EMNIST & Type-1 pixel-pattern & 60 &\textbf{427}(50\%) & 40(50\%) & 336(50\%) & 193(50\%) \\
CIFAR10 & Type-1 pixel-pattern & 40 &\textbf{1501}(50\%) & 837(50\%) & 1068(50\%) & 912(50\%) \\
CIFAR10 & Type-2 pixel-pattern $p_d=0.1$ & 40 &\textbf{833}(50\%) & 358(50\%) & 664(50\%) & 646(50\%) \\
CIFAR10 & Type-2 pixel-pattern $p_d=0.5$ & 40 &\textbf{720}(50\%) & 498(50\%) & 545(50\%) & 438(50\%) \\
CIFAR10 & car with striped bg & 230 &\textbf{1106}(50\%) & 283(50\%) & 216(50\%) & 205(50\%) \\
CIFAR10 & green car & 230 &\textbf{264}(50\%) & 20(50\%) & 76(50\%) & 74(50\%) \\
CIFAR10 & racing car & 230 &\textbf{208}(50\%) & 5(50\%) & 40(50\%) & 70(50\%) \\
CIFAR10 & southwest airline (edge-case) & 230 &339(60\%) & \textbf{349}(60\%) & 75(60\%) & 336(60\%) \\
CIFAR100 & truck with striped bg & 100 &\textbf{625}(60\%) & 623(60\%) & 297(60\%) & 170(60\%) \\
CIFAR100 & bicycle with black bg & 150 &\textbf{661}(50\%) & 229(50\%) & 112(50\%) & 133(50\%) \\
\bottomrule
\end{tabular}
\end{small}
\label{table2}
\end{center}
\vskip -0.1in
\end{table}

\vspace{-3mm}

\begin{table}[h!]
\caption{Lifespan $L(\gamma)$ of the backdoor inserted by Chameleon, Anticipate, Neurotoxin and baseline with ResNet34 model architecture in this paper.}
\begin{center}
\begin{small}
\begin{tabular}{lcccccr}
\toprule
Datasets & Backdoor & AttackNum & Chameleon & Anticipate & Neurotoxin & Baseline \\
\midrule
CIFAR10 & racing car & 230 & \textbf{567}(40\%) & 355(40\%) & 7(40\%) & 13(40\%) \\
\bottomrule
\end{tabular}
\end{small}
\label{table3}
\end{center}
\vskip -0.1in
\end{table}

We present the rest of the experiment results in \cref{figure9}, \cref{figure10}, \cref{figure11}, and summarize the lifespans of all the evaluated tasks in \cref{table2} and \cref{table3}. 

As it is shown in \cref{table2}, lifespans of the backdoors inserted using Chameleon achieve the highest compared to other methods except for the edge-case backdoor where Chameleon achieves a comparable 60\%-Lifespan with Anticipate. Comparing the lifespans of semantic backdoors in CIFAR10 and CIFAR100, we can see that the attacker can plant more durable semantic backdoors into CIFAR100 tasks than CIFAR10 tasks with even less AttackNum required. Also, the edge-case backdoor has stronger durability than other semantic backdoors in CIFAR10 on average. For CIFAR100 tasks, this is because CIFAR100 has more categories in the dataset, which lowers the probability of interferers being selected into the dataset of following benign clients. Following uploaded benign updates will have less conflict with the inserted backdoor, and thus improve durability. Similarly, edge-case backdoor images in CIFAR10 which live on the tail of the input distribution have less in common with interferer images. Thus, future benign updates trained by interferers will also have less conflict with the embedded backdoor, which extends the lifespan of the backdoor compared to other CIFAR10 semantic backdoors. 

\begin{figure}[h!]
\begin{center}
\centerline{\includegraphics{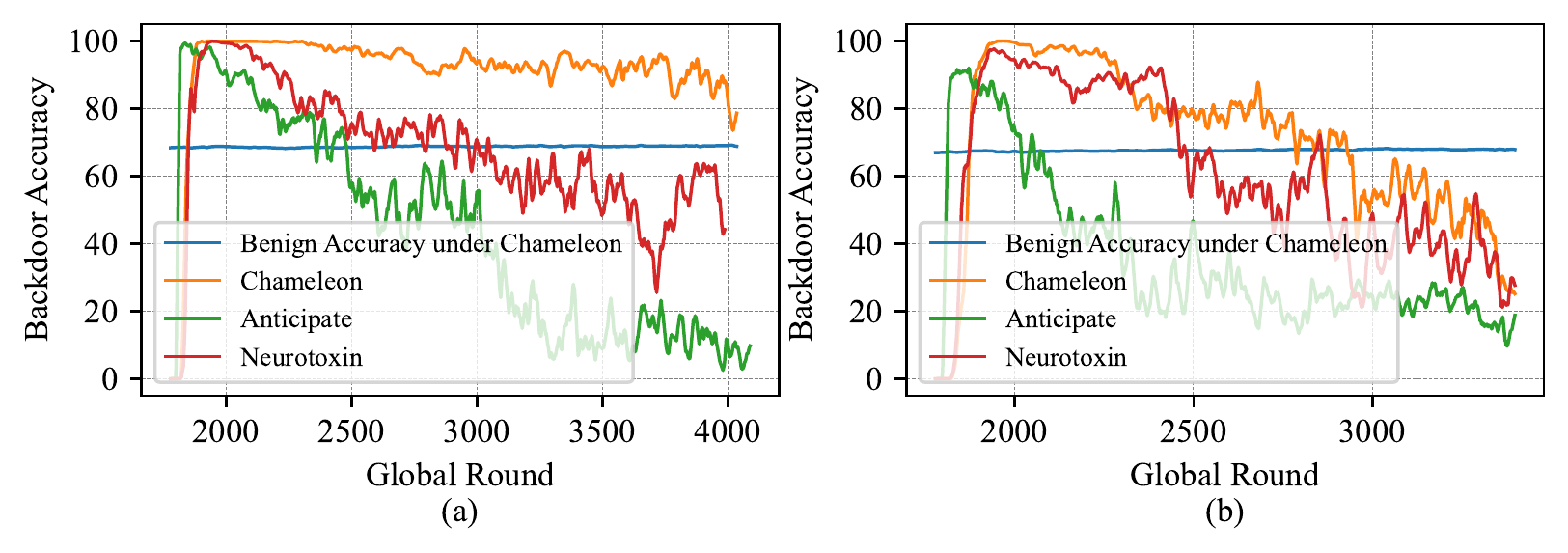}}
\vspace{-3mm}
\caption{Backdoor accuracy when attacking (a) VGG16, (b) VGG19. For all tasks, the attack starts at round 1800, and lasts for AttackNum=150 rounds. The evaluated backdoor is bicycle-with-black-background.}
\label{figure_vgg}
\end{center}
\vskip -0.1in
\end{figure}

\begin{table}[h!]
\caption{50\%-Lifespan of the backdoor inserted by Chameleon, Anticipate and Neurotoxin with VGG16 and VGG19.}
\begin{center}
\begin{small}
\begin{tabular}{cccc}
\toprule
Model & Chameleon & Anticipate & Neurotoxin \\
\midrule
VGG16 & \textbf{2050} & 237 & 530 \\
VGG19 & \textbf{731} & 26 & 489 \\
\bottomrule
\end{tabular}
\end{small}
\label{tabel_vgg}
\end{center}
\vskip -0.1in
\end{table}

We further provide more experimental results to verify that Chameleon is less sensitive to the changes of model architectures. \cref{figure_vgg} and \cref{tabel_vgg} show that the durability of the backdoor inserted by Chameleon achieves the strongest durability when attacking VGG16 or VGG19, which further shows that our method performs well under a wide range of model architectures.

\begin{figure}[ht]
\vskip -0.1in
\begin{center}
\centerline{\includegraphics{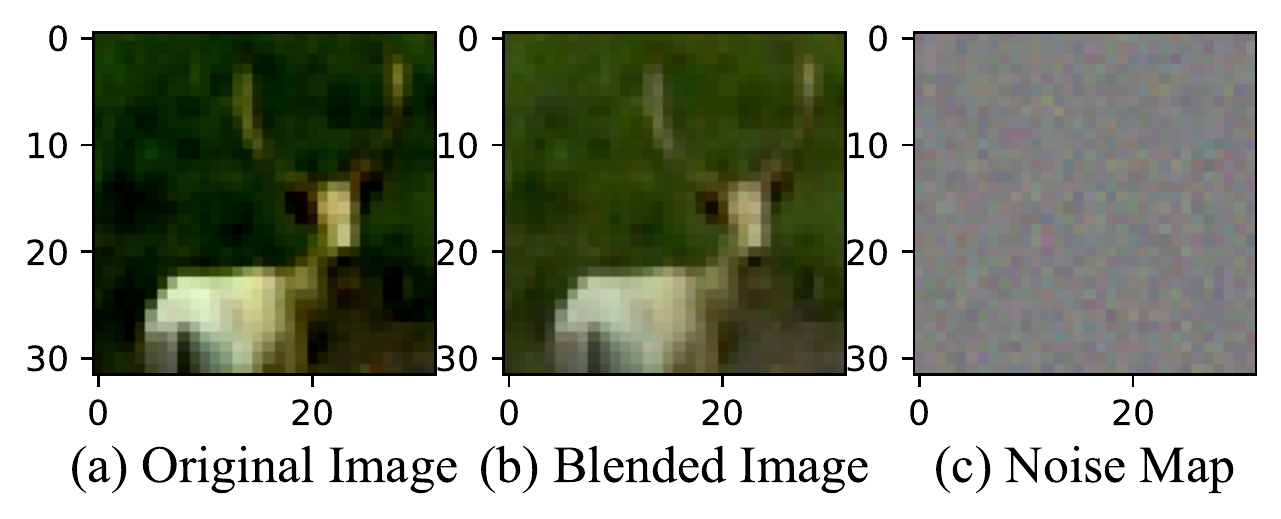}}
\vspace{-3mm}
\caption{(a) Original image. (b) Blended image which is the composition of the original image and the noise map. (c) Noise map which is also the backdoor trigger. The noise map is fixed during the entire process.}
\label{figure_new}
\end{center}
\vskip -0.35in
\end{figure}

%Through carefully designed experiments, we can see that Chameleon has the best performance in extending the lifespan of the inserted backdoor compared to prior methods. And our method is especially effective when there are fewer categories in the dataset and the durability of the inserted backdoor suffers from the update conflicts with interferers and catastrophe forgetting.

%%%%%%%%%%%%%%%%%%%%%%%%%%%%%%%%%%%%%%%%%%%%%%%%%%%%%%%%%%%%%%%%%%%%%%%%%%%%%%%
%%%%%%%%%%%%%%%%%%%%%%%%%%%%%%%%%%%%%%%%%%%%%%%%%%%%%%%%%%%%%%%%%%%%%%%%%%%%%%%

\end{document}